\documentclass{article} 
\usepackage[final]{colm2026_conference}

\usepackage{microtype}
\usepackage{graphicx}
\usepackage{subcaption}
\usepackage{booktabs}
\usepackage{hyperref}
\usepackage{amsmath}
\usepackage[capitalize,noabbrev]{cleveref}
\usepackage{amssymb}
\usepackage{mathtools}
\usepackage{amsthm}
\usepackage{url}
\usepackage{xcolor}
\usepackage{float}
\usepackage{pgfplots}
\usepgfplotslibrary{groupplots}
\usetikzlibrary{pgfplots.groupplots}
\usepackage{longtable}
\usepackage{adjustbox}
\usepackage{lineno}
\usepackage{float}

\usepackage{amsmath,amsfonts,bm}

\def\eqref#1{equation~\ref{#1}}

\def\1{\bm{1}}

\DeclareMathAlphabet{\mathsfit}{\encodingdefault}{\sfdefault}{m}{sl}
\SetMathAlphabet{\mathsfit}{bold}{\encodingdefault}{\sfdefault}{bx}{n}

\definecolor{darkblue}{rgb}{0, 0, 0.5}
\hypersetup{colorlinks=true, citecolor=darkblue, linkcolor=darkblue, urlcolor=darkblue}

\pgfplotsset{
compat=1.18,
seriesA/.style={color=green!60!black, mark=*, mark options={solid}, thin, mark size=1.6pt},
seriesB/.style={color=orange!85!black, mark=square*, mark options={solid}, thin, mark size=1.4pt},
seriesC/.style={color=blue!70!black, mark=triangle*, mark options={solid}, thin, mark size=1.6pt},
seriesD/.style={color=orange!80!black, mark=*, mark options={solid}, thin, mark size=1.6pt},
seriesE/.style={color=cyan!70!blue, mark=diamond*, mark options={solid}, thin, mark size=1.6pt},
legend style={draw=none}
}

\newcommand{\mcnote}[1]{}

\definecolor{olmoeDarkYellow}{HTML}{fdac15}
\definecolor{defaultblue}{HTML}{0077B6}
\definecolor{defaultlightblue}{HTML}{00B4D8}
\definecolor{midblue}{HTML}{0096C7}
\definecolor{blue}{HTML}{03045E}
\definecolor{blueb}{HTML}{0077B6}
\definecolor{bluec}{HTML}{00B4D8}
\definecolor{blued}{HTML}{90E0EF}
\definecolor{bluee}{HTML}{CAF0F8}

\newcommand{\defaultblue}[1]{{\leavevmode\color{blueb}#1}}
\newcommand{\defaultlightblue}[1]{{\leavevmode\color{bluec}#1}}
\newcommand{\defaultyellow}[1]{{\leavevmode\color{olmoeDarkYellow}#1}}

\title{When Fewer Layers Break More Chains: Layer Pruning Harms Test-Time Scaling in LLMs}

\author{
Keyu Wang$^{1,2,3}$\thanks{Equal contribution.} \quad Tian Lyu$^{4}$\footnotemark[1] \quad Guinan Su$^{1,2,3}$ \quad Lu Yin$^{5}$ \quad Marco Canini$^{4}$ \\ \ \textbf{Jonas Geiping}$^{1,2,3}$ \quad \textbf{Shiwei Liu}$^{1,2,3}$ \\
  \ $^{1}$Max Planck Institute for Intelligent Systems \quad
  $^{2}$ELLIS Institute Tübingen \\
  \ $^{3}$Tübingen AI Center \quad
  $^{4}$King Abdullah University of Science and Technology \\
  \ $^{5}$University of Surrey \\
  \  \texttt{keyu.wang@tuebingen.mpg.de} \quad \texttt{sliu@tue.ellis.eu}
}

\begin{document}

\ifcolmsubmission
\linenumbers
\fi

\maketitle

\begin{abstract}
Layer pruning has emerged as a widely adopted technique for improving the efficiency of large language models (LLMs). Although multiple recent works report little-to-no accuracy loss on general knowledge tasks, their effect on long-chain reasoning, a more brittle yet crucial capability, remains largely unexplored. Overlooking this can yield deceptively ``lossless pruning'' that causes catastrophic failures on real-world reasoning workloads. In this work, we study the impact of layer pruning on long-chain reasoning through the lens of test-time scaling, a key mechanism in modern LLMs that enables strong reasoning capacity by allocating more computation at inference time. With extensive experiments, we demonstrate that pruning even one or two layers can severely impair test-time scaling, with performance collapsing drastically on long reasoning benchmarks even when performance on knowledge-intensive tasks remains stable. Furthermore, we find that supervised fine-tuning remedies fail to recover test-time scaling once it has deteriorated. Through in-depth analyses, we identify the mechanisms underlying this fragility of test-time scaling and highlight the fundamental risks of applying layer pruning to reasoning-intensive LLMs. Our findings challenge the prevailing ``lossless pruning'' narrative, call for more sensitive evaluation protocols and cautious deployment in practice. We open-source the codebase in \href{https://github.com/keyu-wang-2002/Layer-Pruning-Harms-Inference-Scaling}{https://github.com/keyu-wang-2002/Layer-Pruning-Harms-Inference-Scaling}. 
\end{abstract}

\section{Introduction}

Recent studies have revealed that large language models exhibit considerable redundancy across their depth, with a substantial fraction of layers contributing minimally to overall functionality \citep{lns, shortgpt, song2026demystifying}. These findings have motivated growing interest in layer pruning, a family of techniques that remove layers to improve efficiency while preserving competitive accuracy \citep{compact, gptailor}. Recent advances highlight its promise: for instance, ShortGPT retains up to 85\% of the original model accuracy after pruning $20\text{--}30\%$ of layers \citep{shortgpt}, and recovers over 90\% with supervised fine-tuning \citep{reassessing}. These results have been widely promoted as evidence that layer pruning is a broadly safe, largely ``lossless'' acceleration strategy. This view has since been widely circulated \citep{shortenedllama, blockpruner, reassessing}.

Despite these encouraging results, prior evaluations of layer pruning have relied primarily on coarse, low-sensitivity benchmarks such as MMLU \citep{mmlu}, which emphasize \textbf{factual knowledge} and \textbf{multiple-choice question}. However, such evaluations overlook long-chain reasoning, a capability central to many LLM applications, such as mathematical problem solving \citep{math500, AIME24}, scientific reasoning \citep{gpqa} and multi-step logical inference \citep{bbh}. Importantly, long-chain reasoning is often more brittle under layer pruning within LLMs. As shown in Figure \ref{intro_fig}, on general knowledge benchmarks such as MMLU, performance of Qwen3-8B degrades only gradually as pruning depth increases by ShortGPT, whereas on tasks requiring long reasoning chains with thousands of tokens such as AIME24 \citep{AIME24}, performance collapses after pruning a single layer and drops to near zero once $10\%$ of layers are removed. This sharp discrepancy suggests that apparent stability on prior standard benchmarks can be a potentially misleading indicator, motivating the need for more sensitive evaluations on long-chain reasoning.

\begin{figure*}[t]
\centering
\begin{tikzpicture}
\begin{groupplot}[
  group style={group size=4 by 1, horizontal sep=0.95cm},
  width=0.267\textwidth, height=0.25\textwidth,
  xtick={0,1,2,3,4,5,6,7},
  grid=major, grid style={dashed,gray!30},
  tick style={line width=0.6pt},
  label style={font=\small, fill=none},
  tick label style={font=\small, fill=none},
  title style={font=\bfseries\small, fill=none},
  axis background/.style={fill=none, draw=none},
  ymin=0, ymax=1
]

\nextgroupplot[title={MMLU}, ylabel={Accuracy}]
\addplot+[seriesE, mark=*] coordinates
  {(0,0.730) (1,0.713) (2,0.661) (3,0.644) (4,0.551) (5,0.538) (6,0.475) (7,0.461)};

\nextgroupplot[title={GSM8K}, ylabel={}]
\addplot+[seriesE, mark=*] coordinates
  {(0,0.879) (1,0.878) (2,0.854) (3,0.712) (4,0.484) (5,0.230) (6,0.002) (7,0.003)};

\nextgroupplot[title={MATH500}, ylabel={}]
\addplot+[seriesE, mark=*] coordinates
  {(0,0.946) (1,0.908) (2,0.780) (3,0.426) (4,0.198) (5,0.186) (6,0.1) (7,0.09)};

\nextgroupplot[title={AIME24}, ylabel={}]
\addplot+[seriesE, mark=*] coordinates
  {(0,0.700) (1,0.467) (2,0.067) (3,0.000) (4,0.000) (5,0.000) (6,0.000) (7,0.000)};

\end{groupplot}

\node[anchor=north,font=\normalsize,yshift=-0.4cm] 
  at ($(group c1r1.south)!0.5!(group c4r1.south)$) {$\#$ of pruned layers by ShortGPT};

\end{tikzpicture}
\caption{Accuracy of the pruned Qwen3-8B by ShortGPT across various pruning depths on MMLU, GSM8K, MATH500 and AIME24. We follow default settings in \textit{lm-eval-harness} \citep{harness}  and additionally apply \textit{LLM-as-a-Judge} to GSM8K, MATH500 and AIME24. }
\label{intro_fig}
\end{figure*}

To study this fragility more systematically and mechanistically, we turn to test-time scaling, a recently emerging technique that provides a finer-grained understanding of how LLMs acquire and exercise long reasoning abilities \citep{s1, demystifying}. By allocating more computation at inference time, such as generating longer chains of thought or exploring multiple reasoning paths \citep{cot, scalinglaws}, test-time scaling has emerged as a key mechanism for enabling human-level reasoning \citep{deepseek-r1}. Crucially, test-time scaling allows us to measure how reasoning ability depends on inference-time computation \citep{tts, s1}, thereby exposing degradations during reasoning process after layer pruning , which standard knowledge or short-context benchmarks fail to capture. Thus, it provides a natural probe for assessing the vulnerability of complex and long reasoning under layer pruning.

In this work, we systematically investigate how layer pruning degrades test-time scaling, covering both sequential scaling (longer chain-of-thought steps) and parallel scaling (more sampled trajectories). Our findings reveal that \textbf{pruning even one or two layers can severely impair test-time scaling}, leading to a catastrophic breakdown in long-chain reasoning capability. We further investigate whether commonly used  supervised fine-tuning remedies can restore the lost scaling behavior, including LoRA fine-tuning \citep{lora} and full-parameter fine-tuning on $1k/10k$-sample reasoning datasets. Extensive experiments demonstrate that \textbf{once degraded, scaling behavior is not effectively restored by supervised fine-tuning.} Finally, we provide in-depth analyses to uncover the underlying reasons, indicating that \textbf{layer pruning causes recurring loops, reduced trajectory diversity and diminished self-reflection} during reasoning and adjusting sampling strategies (temperature, penalty-based sampling) fail to alleviate these problems, pointing to structural damage rather than surface-level accuracy loss. Overall, this work serves as a cautionary evaluation: it challenges the rapid spreading yet potentially misleading assumption that layer pruning is a generally safe ``free lunch'' for LLMs, and clarifies when and why this belief breaks down.

\section{Related Work}

\textbf{LLM Layer Pruning.} Layer pruning aims to reduce the depth of LLMs to improve efficiency while retaining performance. For instance, \citet{shortenedllama} compare pruning strategies based on criteria such as magnitude, perplexity, and Taylor expansion. ShortGPT \citep{shortgpt} highlights the redundancy of LLM layers and introduces Block Influence (BI) to estimate each layer's contribution and prune layers with low BI scores. Similarly, SLEB \citep{sleb} indicates LLMs' redundancy with high similarity between the outputs of neighboring blocks and applies this similarity to eliminate redundant blocks. \citet{reassessing} further shows that parameter-efficient fine-tuning can effectively restore pruned models and \citet{selfdistill} proposes to use self-data distillation for recovering pruned models. Additional analyses of these pruning strategies are presented in \citep{depth, compact}. There also exist structured pruning methods, such as BlockPruner \citep{blockpruner}, LLM-Pruner \citep{llm-pruner}, SliceGPT \citep{slicegpt}, and FinerCut \citep{finercut}, which operate at finer granularities such as channels, attention layers, and feed-forward networks (FFNs). In contrast, our work focuses on pruning at the Transformer layer level. Recently, another line of work proposes merging-based pruning, which fuses adjacent layers to decrease depth without retraining from scratch. Representative methods include MKA \citep{mka}, which aligns layer manifolds to guide merging, and LaCo \citep{laco}, which collapses subsequent layers into a prior layer while preserving their functional capacity. 

\textbf{Curse of Depth.} Recent evidence suggests that modern Pre-LN LLMs increasingly suffer from layer redundancy as depth grows. This phenomenon, known as the Curse of Depth (CoD) \citep{lns}, arises from exponential growth of variance, which drives deeper layers toward near-identity mappings. Existing approaches mitigate CoD by controlling output variance, for example, through scaled initialization \citep{zhang2019improving, takase2023spike}, in-time scaling \citep{lns,dey2026don}, alternative normalization schemes \citep{li2024mix,chen2026post,trochelmann2026kitenorm}, and various types of sparsity \citep{muhtar2026does}. 

\textbf{Test-time Scaling.} Test-time scaling, also called inference scaling, refers to techniques that enhance LLM performance by allocating additional computation at inference time. Recent studies have shown that LLMs benefit significantly from test-time scaling in long and complex reasoning \citep{cot, scalinglaws, deepseek-r1, internvl, kimi, demystifying}. Current advances on test-time scaling mainly fall into two categories: sequential scaling and parallel scaling \citep{ttrl}. Sequential scaling focuses on extending LLM outputs into longer responses by increasing the length of reasoning chains, often through reflective or chain-of-thought processes. For example, s1 \citep{s1} proposes a simple recipe combining a small curated reasoning set and budget forcing, showing consistent gains from longer inference. Similarly, T1 \citep{t1} advances reasoning via reinforcement learning and inference scaling, demonstrating reliable improvements with increased test-time budgets. Concurrently, parallel scaling involves producing multiple candidate responses during inference, either by increasing the number of sampled outputs \citep{ps1} or expanding steps of search \citep{ps2}. The resulting candidates are then integrated through an aggregation strategy, often guided by reward or scoring models \citep{ps4}.

\section{Unveiling the Fragility of Test-Time Scaling under Layer Pruning} \label{section3}

Recent advances in layer pruning have demonstrated that substantial reductions in model depth, e.g., removing up to $20\text{--}30\%$ of layers, can be achieved with only marginal degradation on surface-level metrics such as perplexity or zero-shot accuracy \citep{shortgpt, reassessing}. However, these metrics fail to capture test-time scaling, a critical mechanism for eliciting reasoning capabilities \citep{cot, s1} which is far more brittle and may be sensitive to compression. In this section, we investigate how layer pruning disrupts this scaling capacity.

\textbf{Problem formulation.} In this work, we focus on pruning Transformer layers. Following the notions in \citep{reassessing}, we consider an LLM $\mathcal{M}$ composed of a sequence of Transformer layers $\mathcal{M} = l_1 \circ l_2 \circ \cdots \circ l_n$, where each layer $l_i$ includes a self-attention module and FFNs \citep{transformer}. The objective of layer pruning is to identify a set of layers $\{ l_1^{\prime}, l_2^{\prime}, \dots, l_m^{\prime} \}$ where $m < n$, such that the pruned model $\mathcal{M}'= l_1^{\prime} \circ l_2^{\prime} \circ \cdots \circ l_m^{\prime}$ retains sufficient performance while lowering computational burden.

\textbf{Models.} We study two reasoning models: s1.1-7B \citep{s1} and Qwen3-8B \citep{qwen3}. s1.1-7B is a 7-billion parameter language model finetuned from Qwen2.5-7B-Instruct \citep{qwen2.5} on s1K-1.1 dataset \citep{s1}, which augments reasoning traces with budget forcing to elicit test-time scaling. Qwen3-8B is an 8-billion parameter instruction-tuned model which supports both thinking and non-thinking modes, where the former allocates additional test-time compute for thinking while the latter directly produces concise responses. Both models have been shown to exhibit robust test-time scaling behavior, therefore providing an appropriate testbed for analyzing how layer pruning impacts such reasoning scalability. 

\textbf{Layer pruning methods.} We focus on training-free pruning techniques that reduce model depth while preserving efficiency. For direct removal, we consider (i) \textbf{ShortGPT} \citep{shortgpt}, which eliminates layers with low Block Influence scores, and (ii) \textbf{Reverse-order} \citep{shortenedllama}, which removes deeper layers. For merging-based pruning, we adopt (iii) \textbf{LaCo} \citep{laco}, which linearly combines adjacent layers to preserve representational capacity. Appendix \ref{Appendix_pruning} provides their detailed explanations and implementation specifics in this work. We select these three methods for their strong empirical performance and complementary strategies, direct removal versus layer merging, providing diverse perspectives on how layer pruning influences test-time scaling.

\textbf{Evaluation dimensions.} We evaluate the effect of layer pruning on test-time scaling along two orthogonal dimensions: (1) \textbf{Sequential scaling.} We measure performance under reasoning with increasing thinking token budgets of $[512, 1024, 2048, 4096, 8192]$. We set the temperature to $1.0$ and run experiments with three randomly selected seeds, reporting the average results. This setting reflects iterative reasoning, where later computations build upon earlier steps, allowing reflective and chain-of-thought steps into longer reasoning chains. (2) \textbf{Parallel scaling.} We report $pass@k$ with temperature $1.0$, which means the probability that at least one correct solution exists among $k$ randomly sampled outputs, with $k = [1, 2, 4, 8, 16, 32]$. This setting captures parallel inference, where multiple reasoning trajectories are sampled independently and the results are aggregated.

\textbf{Evaluation datasets and metrics.} We evaluate on three standard reasoning benchmarks: (1) \textbf{MATH500} \citep{math500}, 500 competition-level math problems spanning diverse topics and difficulty levels; (2) \textbf{GPQA Diamond} \citep{gpqa}, 198 graduate-level biology, physics, and chemistry questions targeting expert scientific reasoning; (3) \textbf{AIME24} \citep{AIME24}, 30 problems from the 2024 American Invitational Mathematics Examination, covering algebra, geometry, number theory, and probability. Our evaluation framework builds on \textit{lm-evaluation-harness} \citep{harness} and adopts \textit{LLM-as-a-Judge} to avoid influence from the output format \citep{s1}.

\begin{figure*}[t]
\centering
\begin{tikzpicture}
\begin{groupplot}[
  group style={group size=3 by 3, horizontal sep=1.2cm, vertical sep=0.7cm},
  width=0.31\textwidth, height=0.23\textwidth,
  xmode=log, log basis x=2,
  xtick={512,1024,2048,4096,8192},
  xticklabels={512,1024,2048,4096,8192},
  grid=major, grid style={dashed,gray!30},
  tick style={line width=0.6pt},
  label style={font=\small, fill=none},
  tick label style={font=\small, fill=none},
  xticklabel style={font=\scriptsize},
  title style={font=\bfseries\small, fill=none},
  axis background/.style={fill=none, draw=none},
  ylabel={Accuracy},
]

\nextgroupplot[
  title={MATH500},
  ylabel={Acc (ShortGPT)},
  legend to name=legendQwenScalingShortGPT,
  legend columns=4,
  legend style={draw=none, /tikz/every even column/.style={column sep=0.8em}, font=\small}
]
\addplot+[seriesA] coordinates {(512,0.875) (1024,0.905) (2048,0.924) (4096,0.929) (8192,0.960)};
\addlegendentry{No pruning}
\addplot+[seriesB] coordinates {(512,0.811) (1024,0.837) (2048,0.863) (4096,0.892) (8192,0.917)};
\addlegendentry{Prune 1 layer}
\addplot+[seriesC] coordinates {(512,0.751) (1024,0.762) (2048,0.753) (4096,0.781) (8192,0.787)};
\addlegendentry{Prune 2 layers}
\addplot+[seriesE] coordinates {(512,0.342) (1024,0.384) (2048,0.42) (4096,0.438) (8192,0.434)};
\addlegendentry{Prune 3 layers}

\nextgroupplot[title={GPQA Diamond}, ylabel={}]
\addplot+[seriesA] coordinates {(512,0.446) (1024,0.441) (2048,0.478) (4096,0.530) (8192,0.564)};
\addplot+[seriesB] coordinates {(512,0.416) (1024,0.407) (2048,0.432) (4096,0.488) (8192,0.528)};
\addplot+[seriesC] coordinates {(512,0.411) (1024,0.379) (2048,0.392) (4096,0.402) (8192,0.421)};
\addplot+[seriesE] coordinates {(512,0.369) (1024,0.348) (2048,0.364) (4096,0.333) (8192,0.323)};

\nextgroupplot[
  title={AIME24}, ylabel={},
  scaled y ticks=false,
  y tick label style={/pgf/number format/fixed,/pgf/number format/precision=2}
]
\addplot+[seriesA] coordinates {(512,0.556) (1024,0.422) (2048,0.489) (4096,0.467) (8192,0.667)};
\addplot+[seriesB] coordinates {(512,0.367) (1024,0.244) (2048,0.333) (4096,0.378) (8192,0.467)};
\addplot+[seriesC] coordinates {(512,0.100) (1024,0.133) (2048,0.067) (4096,0.100) (8192,0.156)};
\addplot+[seriesE] coordinates {(512,0.067) (1024,0.067) (2048,0.0) (4096,0.0) (8192,0.067)};

\nextgroupplot[
  ylabel={Acc (Reverse)},
  legend to name=legendQwenScalingReverse,
  legend columns=4,
  legend style={draw=none, /tikz/every even column/.style={column sep=0.8em}, font=\small}
]
\addplot+[seriesA] coordinates {(512,0.875) (1024,0.905) (2048,0.924) (4096,0.929) (8192,0.960)};
\addlegendentry{No pruning}
\addplot+[seriesB] coordinates {(512,0.725) (1024,0.781) (2048,0.924) (4096,0.944) (8192,0.936)};
\addlegendentry{Prune 1 layer}
\addplot+[seriesC] coordinates {(512,0.765) (1024,0.789) (2048,0.776) (4096,0.786) (8192,0.780)};
\addlegendentry{Prune 2 layers}
\addplot+[seriesE] coordinates {(512,0.132) (1024,0.162) (2048,0.12) (4096,0.196) (8192,0.182)};
\addlegendentry{Prune 3 layers}

\nextgroupplot[ylabel={}]
\addplot+[seriesA] coordinates {(512,0.446) (1024,0.441) (2048,0.478) (4096,0.530) (8192,0.564)};
\addplot+[seriesB] coordinates {(512,0.439) (1024,0.449) (2048,0.479) (4096,0.522) (8192,0.551)};
\addplot+[seriesC] coordinates {(512,0.422) (1024,0.404) (2048,0.414) (4096,0.422) (8192,0.432)};
\addplot+[seriesE] coordinates {(512,0.373) (1024,0.373) (2048,0.373) (4096,0.368) (8192,0.394)};

\nextgroupplot[
  ylabel={}, scaled y ticks=false,
  y tick label style={/pgf/number format/fixed,/pgf/number format/precision=2}
]
\addplot+[seriesA] coordinates {(512,0.556) (1024,0.422) (2048,0.489) (4096,0.467) (8192,0.667)};
\addplot+[seriesB] coordinates {(512,0.400) (1024,0.366) (2048,0.150) (4096,0.217) (8192,0.283)};
\addplot+[seriesC] coordinates {(512,0.067) (1024,0.100) (2048,0.167) (4096,0.200) (8192,0.212)};
\addplot+[seriesE] coordinates {(512,0.0) (1024,0.0) (2048,0.0) (4096,0.0) (8192,0.0)};
\nextgroupplot[
  ylabel={Acc (LaCo)},
  xlabel={Thinking tokens},
  legend to name=legendQwenScalingLaCo,
  legend columns=3,
  legend style={draw=none, /tikz/every even column/.style={column sep=0.8em}, font=\small}
]
\addplot+[seriesA] coordinates {(512,0.875) (1024,0.905) (2048,0.924) (4096,0.929) (8192,0.960)};
\addlegendentry{No pruning}
\addplot+[seriesB] coordinates {(512,0.788) (1024,0.818) (2048,0.862) (4096,0.878) (8192,0.874)};
\addlegendentry{Prune 1 layer}
\addplot+[seriesC] coordinates {(512,0.693) (1024,0.763) (2048,0.770) (4096,0.824) (8192,0.829)};
\addlegendentry{Prune 2 layers}
\addplot+[seriesE] coordinates {(512,0.559) (1024,0.596) (2048,0.648) (4096,0.643) (8192,0.619)};
\addlegendentry{Prune 3 layers}

\nextgroupplot[ylabel={}, xlabel={Thinking tokens}]
\addplot+[seriesA] coordinates {(512,0.446) (1024,0.441) (2048,0.478) (4096,0.530) (8192,0.564)};
\addplot+[seriesB] coordinates {(512,0.448) (1024,0.470) (2048,0.488) (4096,0.565) (8192,0.614)};
\addplot+[seriesC] coordinates {(512,0.372) (1024,0.374) (2048,0.366) (4096,0.404) (8192,0.414)};
\addplot+[seriesE] coordinates {(512,0.389) (1024,0.358) (2048,0.429) (4096,0.434) (8192,0.444)};

\nextgroupplot[
  ylabel={}, xlabel={Thinking tokens},
  scaled y ticks=false,
  y tick label style={/pgf/number format/fixed,/pgf/number format/precision=2}
]
\addplot+[seriesA] coordinates {(512,0.556) (1024,0.422) (2048,0.489) (4096,0.467) (8192,0.667)};
\addplot+[seriesB] coordinates {(512,0.411) (1024,0.367) (2048,0.356) (4096,0.467) (8192,0.522)};
\addplot+[seriesC] coordinates {(512,0.178) (1024,0.267) (2048,0.234) (4096,0.256) (8192,0.234)};
\addplot+[seriesE] coordinates {(512,0.055) (1024,0.011) (2048,0.022) (4096,0.178) (8192,0.211)};

\end{groupplot}
\end{tikzpicture}

\vspace{0.4em}
\pgfplotslegendfromname{legendQwenScalingShortGPT}

\caption{Sequential test-time scaling of Qwen3-8B under different pruning depths.}

\label{fig-qwen-scaling}
\end{figure*}
\begin{figure*}[t]
\centering
\begin{tikzpicture}
\begin{groupplot}[
  group style={group size=3 by 1, horizontal sep=1.2cm, vertical sep=0.0cm},
  width=0.31\textwidth, height=0.23\textwidth,
  xmode=log, log basis x=2,
  xtick={1, 2,4,8,16,32},
  xticklabels={1, 2,4,8,16,32},
  grid=major, grid style={dashed,gray!30},
  tick style={line width=0.6pt},
  label style={font=\small, fill=none},
  tick label style={font=\small, fill=none},
  title style={font=\bfseries\small, fill=none},
  axis background/.style={fill=none, draw=none},
  ylabel={Pass$@$k on AIME24},
  xmin=1, xmax=32,
  enlarge x limits=0.05,
  enlarge y limits=0.1
]

\nextgroupplot[
  title={ShortGPT},
  xlabel={$k$},
  legend to name=legendQwenPassKShortGPT,
  legend columns=4,
  legend style={draw=none, /tikz/every even column/.style={column sep=0.8em}, font=\small}
]
\addplot+[seriesA] coordinates {(1,0.667) (2,0.784) (4,0.818) (8,0.842) (16,0.865) (32,0.885)};
\addlegendentry{No pruning}
\addplot+[seriesB] coordinates {(1,0.467) (2,0.542) (4,0.626) (8,0.698) (16,0.763) (32,0.822)};
\addlegendentry{Prune 1 layer}
\addplot+[seriesC] coordinates {(1,0.156) (2,0.171) (4,0.210) (8,0.252) (16,0.301) (32,0.344)};
\addlegendentry{Prune 2 layers}
\addplot+[seriesE] coordinates {(1,0.067) (2,0.059) (4,0.096) (8,0.147) (16,0.206) (32,0.267)};
\addlegendentry{Prune 3 layers}

\nextgroupplot[
  title={Reverse-order},
  ylabel={},
  xlabel={$k$},
]
\addplot+[seriesA] coordinates {(1,0.667) (2,0.784) (4,0.818) (8,0.842) (16,0.865) (32,0.885)};
\addplot+[seriesB] coordinates {(1,0.467) (2,0.579) (4,0.690) (8,0.762) (16,0.805) (32,0.833)};
\addplot+[seriesC] coordinates {(1, 0.172) (2,0.195) (4,0.239) (8,0.278) (16,0.302) (32,0.346)};
\addplot+[seriesE] coordinates {(1,0.033) (2,0.002) (4,0.004) (8,0.008) (16,0.014) (32,0.022)};

\nextgroupplot[
  title={LaCo},
  ylabel={},
  xlabel={$k$},
]
\addplot+[seriesA] coordinates {(1, 0.667) (2,0.784) (4,0.818) (8,0.842) (16,0.865) (32,0.885)};
\addplot+[seriesB] coordinates {(1, 0.522) (2,0.730) (4,0.797) (8,0.836) (16,0.862) (32,0.882)};
\addplot+[seriesC] coordinates {(1, 0.234) (2,0.306) (4,0.387) (8,0.460) (16,0.522) (32,0.571)};
\addplot+[seriesE] coordinates {(1,0.211) (2,0.270) (4,0.352) (8,0.427) (16,0.488) (32,0.529)};

\end{groupplot}
\end{tikzpicture}

\vspace{0.4em}
\pgfplotslegendfromname{legendQwenPassKShortGPT}

\caption{Parallel test-time scaling of Qwen3-8B on AIME24 under different pruning depths. }
\label{fig-passk-aime24}
\end{figure*}

\textbf{Results.} For sequential scaling, results for Qwen3-8B are reported in Figure~\ref{fig-qwen-scaling}, and those for s1.1-7B are provided in Figure~\ref{fig-s1-scaling} in Appendix~\ref{Appendix_scaling}. Overall, layer pruning severely disrupts sequential test-time scaling. Qwen3-8B is comparatively more resilient than s1.1-7B: after pruning one layer, it largely preserves sequential scaling, with performance continuing to improve as thinking tokens increase, and on GPQA Diamond it even approaches or slightly surpasses the original model. However, pruning two or three layers completely collapses sequential scaling, causing a sharp performance drop. Moreover, as the number of thinking tokens increases, performance does not show significant improvement but instead tends to plateau. Consistent with this trend, s1.1-7B exhibits greater fragility: pruning even a single layer substantially impairs scaling. With two layers pruned, scaling curve vanishes and performance becomes flat or even degrades as thinking tokens increase. Taken together, these results indicate that sequential test-time scaling is highly fragile under layer pruning, breaking down once pruning depth exceeds a very shallow threshold.

For parallel scaling, we report the results of Qwen3-8B in Figure~\ref{fig-passk-aime24}. Overall, layer pruning causes a substantial drop in $Pass@k$ across a wide range of $k$. Because the unpruned Qwen3-8B already achieves strong $Pass@1$ on AIME24, increasing $k$ yields only modest additional gains. After layer pruning, parallel scaling still provides some benefit. Nevertheless, pruning consistently shifts the curves downward via significantly reducing both $Pass@1$ and the attainable ceiling, so that once two or more layers are removed, a substantial gap to the unpruned model remains and cannot be bridged simply by sampling more trajectories.

\section{Recovering Test-Time Scaling: Is SFT Sufficient?} \label{section4}

The fragility of test-time scaling under layer pruning, revealed in Section~\ref{section3}, raises a natural question: \textit{Can supervised fine-tuning (SFT) restore test-time scaling after layer pruning?} While prior work has shown that SFT can substantially improve the performance of pruned LLMs, such evaluations have largely been restricted to relatively simple language and knowledge tasks such as MMLU and HellaSwag \citep{reassessing}. Whether SFT can restore the disrupted test-time scaling remains a non-trivial open problem. In this section, we study two representative strategies, LoRA fine-tuning (LoRA FT) \citep{lora} and full-parameter fine-tuning (Full FT), using SFT datasets of two sizes with 1k and 10k samples, to assess their effectiveness in recovering test-time scaling degraded by layer pruning.

\textbf{Experimental Settings.} Since ShortGPT, Reverse-order, and LaCo all induce more severe degradation of sequential scaling, we employ the same settings used for sequential scaling in Section~\ref{section3}. For SFT settings: (1) \textbf{LoRA FT on 1k data.} We fine-tune the pruned variants of s1.1-7B and Qwen3-8B on the s1K-1.1 dataset \citep{s1} using LoRA FT with rank 16 and scaling factor 32. For each pruned model, we perform a grid search over learning rates $[1{e}{-6}, 4{e}{-6}, 1{e}{-5}, 4{e}{-5}, 1{e}{-4}]$, selecting the best value by validation on MATH500, GPQA Diamond and AIME24, respectively. Results are reported on the corresponding datasets, with detailed learning rate sweeps provided in Appendix~\ref{Appendix_lr}. (2) \textbf{Full FT on 1k and 10k data.} We follow the s1 codebase and configuration \citep{s1}, and perform Full FT with a fixed learning rate of $1e-5$. We use two SFT datasets: (i) \textsc{s1K-1.1}, and (ii) a 10k-sample subset randomly selected from OpenR1-Math-220k\footnote{\url{https://huggingface.co/datasets/open-r1/OpenR1-Math-220k}}. We provide more details about the datasets in Appendix \ref{sec:datasets}.

\begin{figure*}[t]
\centering
\begin{tikzpicture}
\begin{groupplot}[
  group style={group size=3 by 3, horizontal sep=1.2cm, vertical sep=0.7cm},
  width=0.31\textwidth, height=0.23\textwidth,
  xmode=log, log basis x=2,
  xtick={512,1024,2048,4096,8192},
  xticklabels={512,1024,2048,4096,8192},
  grid=major, grid style={dashed,gray!30},
  tick style={line width=0.6pt},
  label style={font=\small, fill=none},
  tick label style={font=\small, fill=none},
  xticklabel style={font=\scriptsize},
  title style={font=\bfseries\small, fill=none},
  axis background/.style={fill=none, draw=none},
  ylabel={Accuracy},
]

\nextgroupplot[
  title={MATH500},
  ylabel={Acc (ShortGPT)},
  legend to name=legendQwenSftTenKShortGPT,
  legend columns=4,
  legend style={draw=none, /tikz/every even column/.style={column sep=0.8em}, font=\small}
]
\addplot+[seriesA] coordinates {(512,0.875) (1024,0.905) (2048,0.924) (4096,0.929) (8192,0.960)};
\addlegendentry{No pruning}
\addplot+[seriesB] coordinates {(512,0.721) (1024,0.791) (2048,0.819) (4096,0.818) (8192,0.826)};
\addlegendentry{Prune 1 layer}
\addplot+[seriesC] coordinates {(512,0.744) (1024,0.813) (2048,0.819) (4096,0.810) (8192,0.835)};
\addlegendentry{Prune 2 layers}
\addplot+[seriesE] coordinates {(512,0.704) (1024,0.753) (2048,0.784) (4096,0.789) (8192,0.797)};
\addlegendentry{Prune 3 layers}

\nextgroupplot[title={GPQA Diamond}, ylabel={}]
\addplot+[seriesA] coordinates {(512,0.446) (1024,0.441) (2048,0.478) (4096,0.530) (8192,0.564)};
\addplot+[seriesB] coordinates {(512,0.326) (1024,0.362) (2048,0.404) (4096,0.414) (8192,0.445)};
\addplot+[seriesC] coordinates {(512,0.368) (1024,0.355) (2048,0.365) (4096,0.373) (8192,0.408)};
\addplot+[seriesE] coordinates {(512,0.362) (1024,0.363) (2048,0.350) (4096,0.354) (8192,0.374)};

\nextgroupplot[
  title={AIME24}, ylabel={},
  scaled y ticks=false,
  y tick label style={/pgf/number format/fixed,/pgf/number format/precision=2}
]
\addplot+[seriesA] coordinates {(512,0.556) (1024,0.422) (2048,0.489) (4096,0.467) (8192,0.667)};
\addplot+[seriesB] coordinates {(512,0.233) (1024,0.267) (2048,0.378) (4096,0.367) (8192,0.445)};
\addplot+[seriesC] coordinates {(512,0.233) (1024,0.189) (2048,0.289) (4096,0.267) (8192,0.244)};
\addplot+[seriesE] coordinates {(512,0.145) (1024,0.155) (2048,0.200) (4096,0.233) (8192,0.189)};

\nextgroupplot[
  ylabel={Acc (Reverse)},
  legend to name=legendQwenSftTenKReverse,
  legend columns=4,
  legend style={draw=none, /tikz/every even column/.style={column sep=0.8em}, font=\small}
]
\addplot+[seriesA] coordinates {(512,0.875) (1024,0.905) (2048,0.924) (4096,0.929) (8192,0.960)};
\addlegendentry{No pruning}
\addplot+[seriesB] coordinates {(512,0.683) (1024,0.808) (2048,0.839) (4096,0.868) (8192,0.856)};
\addlegendentry{Prune 1 layer}
\addplot+[seriesC] coordinates {(512,0.713) (1024,0.817) (2048,0.835) (4096,0.856) (8192,0.850)};
\addlegendentry{Prune 2 layers}
\addplot+[seriesE] coordinates {(512,0.758) (1024,0.839) (2048,0.834) (4096,0.854) (8192,0.838)};
\addlegendentry{Prune 3 layers}

\nextgroupplot[ylabel={}]
\addplot+[seriesA] coordinates {(512,0.446) (1024,0.441) (2048,0.478) (4096,0.530) (8192,0.564)};
\addplot+[seriesB] coordinates {(512,0.372) (1024,0.372) (2048,0.446) (4096,0.443) (8192,0.456)};
\addplot+[seriesC] coordinates {(512,0.404) (1024,0.404) (2048,0.451) (4096,0.429) (8192,0.444)};
\addplot+[seriesE] coordinates {(512,0.392) (1024,0.397) (2048,0.392) (4096,0.461) (8192,0.463)};

\nextgroupplot[
  ylabel={}, scaled y ticks=false,
  y tick label style={/pgf/number format/fixed,/pgf/number format/precision=2}
]
\addplot+[seriesA] coordinates {(512,0.556) (1024,0.422) (2048,0.489) (4096,0.467) (8192,0.667)};
\addplot+[seriesB] coordinates {(512,0.345) (1024,0.289) (2048,0.356) (4096,0.367) (8192,0.467)};
\addplot+[seriesC] coordinates {(512,0.378) (1024,0.267) (2048,0.400) (4096,0.389) (8192,0.400)};
\addplot+[seriesE] coordinates {(512,0.300) (1024,0.333) (2048,0.366) (4096,0.378) (8192,0.411)};

\nextgroupplot[
  ylabel={Acc (LaCo)},
  xlabel={Thinking tokens},
  legend to name=legendQwenSftTenKLaCo,
  legend columns=3,
  legend style={draw=none, /tikz/every even column/.style={column sep=0.8em}, font=\small}
]
\addplot+[seriesA] coordinates {(512,0.875) (1024,0.905) (2048,0.924) (4096,0.929) (8192,0.960)};
\addlegendentry{No pruning}
\addplot+[seriesB] coordinates {(512,0.829) (1024,0.888) (2048,0.882) (4096,0.878) (8192,0.850)};
\addlegendentry{Prune 1 layer}
\addplot+[seriesC] coordinates {(512,0.784) (1024,0.838) (2048,0.847) (4096,0.867) (8192,0.826)};
\addlegendentry{Prune 2 layers}
\addplot+[seriesE] coordinates {(512,0.836) (1024,0.865) (2048,0.838) (4096,0.846) (8192,0.837)};
\addlegendentry{Prune 3 layers}

\nextgroupplot[ylabel={}, xlabel={Thinking tokens}]
\addplot+[seriesA] coordinates {(512,0.446) (1024,0.441) (2048,0.478) (4096,0.530) (8192,0.564)};
\addplot+[seriesB] coordinates {(512,0.495) (1024,0.497) (2048,0.518) (4096,0.480) (8192,0.515)};
\addplot+[seriesC] coordinates {(512,0.401) (1024,0.409) (2048,0.438) (4096,0.459) (8192,0.474)};
\addplot+[seriesE] coordinates {(512,0.449) (1024,0.450) (2048,0.444) (4096,0.431) (8192,0.432)};

\nextgroupplot[
  ylabel={}, xlabel={Thinking tokens},
  scaled y ticks=false,
  y tick label style={/pgf/number format/fixed,/pgf/number format/precision=2}
]
\addplot+[seriesA] coordinates {(512,0.556) (1024,0.422) (2048,0.489) (4096,0.467) (8192,0.667)};
\addplot+[seriesB] coordinates {(512,0.455) (1024,0.445) (2048,0.444) (4096,0.433) (8192,0.400)};
\addplot+[seriesC] coordinates {(512,0.322) (1024,0.322) (2048,0.400) (4096,0.355) (8192,0.300)};
\addplot+[seriesE] coordinates {(512,0.367) (1024,0.400) (2048,0.411) (4096,0.356) (8192,0.378)};

\end{groupplot}
\end{tikzpicture}

\vspace{0.4em}
\pgfplotslegendfromname{legendQwenSftTenKShortGPT}

\caption{Sequential scaling of Qwen3-8B accross pruning depths after Full SFT on 10k data.}
\label{fig-qwen-scaling-sft-10k}
\end{figure*}

\textbf{Results.} Figure~\ref{fig-qwen-scaling-sft-10k} presents the sequential scaling performance of pruned Qwen3-8B after Full FT on 10k data, which achieves the best performance among all the SFT settings. The corresponding results for pruned Qwen3-8B after Full FT and Lora FT on 1k data, pruned s1.1-7B after Full FT and LoRA FT on 1k data, are shown in Figures~\ref{fig:fig-qwen-scaling-merge} and ~\ref{fig:fig-s1-scaling-merge} in Appendix~\ref{Appendix_scaling_ft}, respectively. Overall, SFT on 10k data outperforms SFT on 1k data and Full FT outperforms LoRA FT, but all SFT settings fail to restore the sequential test-time scaling disrupted by layer pruning. For Full FT on 10k samples, the most substantial gains are observed under pruning 2-3 layers, where SFT provides significant improvements, but it still remains far from reaching the original model's performance. Under mild 1-layer pruning, the improvements are limited and can be inconsistent, suggesting that  SFT may emphasize distributional drift rather than delivering meaningful repair. More importantly, as the number of thinking tokens increases, the performance gap to the unpruned model continues to widen, indicating that this SFT fails to restore sequential scaling. Additionally, under SFT on 1k samples, we observe a similar trend to the 10k setting, but with markedly smaller gains. Overall, these results indicate that SFT cannot repair the fundamental breakdown of sequential test-time scaling, suggesting that SFT is unlikely to induce the global shift in weight distributions required to fundamentally restore reasoning scalability. 

These findings do not preclude the possibility that large-scale re-training could recover the lost scaling behavior. However, achieving such a shift would likely require access to the original pre-training data to re-train the model at scale, which is typically not accessible for most deployed LLMs.

\section{Analysis}

\subsection{Repetitive Reasoning Loops after Layer Pruning} \label{Section_5.1}

To better understand why layer pruning severely disrupts test-time scaling, we conduct a detailed case study. As shown in Figure~\ref{fig:examples_table}, layer pruning impairs the model's ability to sustain coherent reasoning trajectories. Instead, the pruned models often fall into \textbf{repetitive self-doubt} and \textbf{looping deduction}. For example, in the MATH500 case, the model initially identifies the correct candidate solution but repeatedly questions itself, rechecking the same invalid negative cases, e.g., \textit{``But wait, let me check again...''}, without making progress. In many cases, the pruned model initially arrives at a promising intermediate step, but rather than building on it, the reasoning derails into repetitive checks of already invalid branches, or into circular speculation that never converges. Once stuck, the model fails to explore alternative directions and keeps generating redundant text without real progress.  By pruning layers, we inadvertently collapse the model's reasoning diversity, causing breakdowns in long chain-of-thought and test-time scaling where longer thinking no longer leads to stronger performance. We provide the complete examples in Appendix~\ref{Appendix_qualitative_example}.

\definecolor{defaultblue}{HTML}{0077B6}
\definecolor{defaultlightblue}{HTML}{00B4D8}
\definecolor{teal}{HTML}{2A9D8F}

\begin{figure*}[t]
\footnotesize
\centering
\begin{tabular}{@{}p{0.31\textwidth}p{0.31\textwidth}p{0.31\textwidth}@{}}
\toprule

\textbf{[MATH500]}

Find the number of integer values of $k \in [-500, 500]$ for which $\log(kx) = 2\log(x+2)$ has exactly one real solution.

~

\textcolor{teal}{...Therefore, the only possible $k$ is $k=8$...}\textcolor{defaultlightblue}{
\textit{ But wait, let me check again... But wait... Let me think. If $k$ is negative... But wait, let me check if there are any other k values... Wait, in Case 2, k is negative. Let me check if for any negative k... Wait, in Case 2, x is negative... But wait, let me check again... But wait, let me check if there are any other k values... Thus, k = 0 or k = 8. But k must be negative, so k = 0 is not valid. Therefore, there is no solution where k is negative... }
}

~

\textcolor{defaultblue}{
Setup is correct, but domain analysis for $k<0$ causes repetitive self-doubt.
}
&
\textbf{[GPQA Diamond]}

Two quantum states with lifetimes $10^{-9}\,\text{s}$ and $10^{-8}\,\text{s}$ need to be clearly resolved. What is a possible energy difference?

~

\textcolor{teal}{The lifetime of a quantum state is related to its uncertainty in energy. The uncertainty principle tells us that $\Delta E \Delta t \geq \hbar/2$...} \textcolor{defaultlightblue}{
 \textit{Wait, but how does the energy difference relate to the lifetimes?... Wait, maybe the question is referring to resolving the two states in time?...  Wait, maybe the question is not about the sum... Wait, no... Wait, maybe the question is about resolving the two states in energy... Wait, the uncertainty principle is...}
}

~

\textcolor{defaultblue}{
 Correct principle cited, but application is inconsistent, leading to  repetitive chains.
}

&

\textbf{[AIME24]} 

Let $p$ be the least prime for which $n^4+1 \equiv 0 \pmod{p^2}$. Find the least positive $m$ such that $m^4+1 \equiv 0 \pmod{p^2}$.

~

\textcolor{teal}{The smallest prime $p$ congruent to 1 modulo 8 is 17... }\textcolor{defaultlightblue}{17 divides 83521-17=83504. Therefore, 17 divides 83521-17, so 17 divides 83521-17. Therefore, 17 divides 83521-17. Therefore, 17 divides 83521-17. Therefore, 17 divides 83521-17. Therefore, 17 divides 83521-17. Therefore, 17 divides 83521-17. Therefore, 17 divides 83521-17. Therefore, 17 divides 83521-17. Therefore, 17 divides 83521-17. Therefore, 17 divides 83521-17. ...
}

~

\textcolor{defaultblue}{
Finds $p=17$ but fails to lift the root modulo $p^2$, leading to reasoning loops.
} 
\\
\bottomrule
\end{tabular}
\vspace{-5pt}
\caption{Examples of repetitive thinking from 1-layer pruned s1.1-7B model. Unlike the original model, the pruned version gets stuck in logic loops. The \textcolor{teal}{teal}  text denotes the correct reasoning trace while \textcolor{defaultlightblue}{light blue} text is the repetitive reasoning trace leading to failure, and the \textcolor{defaultblue}{blue} text is a brief summary of the failure reason. }
\label{fig:examples_table}
\end{figure*}
\begin{table*}[h]
\centering
\scriptsize
\setlength{\tabcolsep}{2.5pt}
\resizebox{\linewidth}{!}{
\begin{tabular}{l|c|cccc|cccc|cccc}
\toprule
& \multicolumn{1}{c|}{\textbf{Original}} 
& \multicolumn{4}{c|}{\textbf{ShortGPT}} 
& \multicolumn{4}{c|}{\textbf{Reverse-order}} 
& \multicolumn{4}{c}{\textbf{LaCo}} \\
\cmidrule(lr){2-2} \cmidrule(lr){3-6} \cmidrule(lr){7-10} \cmidrule(lr){11-14}
\textbf{Model/Method} 
& \textbf{original} 
& \textbf{1 layer} & \textbf{2 layer} & \textbf{1 layer SFT} & \textbf{2 layer SFT} 
& \textbf{1 layer} & \textbf{2 layer} & \textbf{1 layer SFT} & \textbf{2 layer SFT} 
& \textbf{1 layer} & \textbf{2 layer} & \textbf{1 layer SFT} & \textbf{2 layer SFT} \\
\midrule
\textbf{s1.1-7B}  & 0.3 & 0.83 & 0.87 & 0.0 & 0.0 & 0.97 & 0.63 & 0.0 & 0.0 & 0.77 & 1.0 & 0.0 & 0.0 \\
\textbf{Qwen3-8B} & 0.1 & 0.4 & 1.0 & 0.03 & 0.0 & 0.53 & 0.87 & 0.03 & 0.03 & 0.03 & 0.1 & 0.03 & 0.0 \\
\bottomrule

\end{tabular}
}
\caption{Loop fractions for pruned models ($\downarrow$). ``1 layer'' and ``2 layer'' indicate $\#$ of pruned layers, whereas ``SFT'' refer to Full FT.}
\label{tab:loopfraction}
\end{table*}



To quantify these observations, we first employ Loop Fraction~\citep{loopfraction} to measure verbatim repetition. We consider a text to exhibit looping if it contains any $n$-gram that appears at least $k$ times, and set $n=30$ and $k=10$ in our study. Table~\ref{tab:loopfraction} shows that layer pruning significantly increases text loops. For example, the original s1.1-7B exhibits a relatively low loop fraction (0.3), which rises to over 0.8 after pruning one layer with ShortGPT. Similarly, Qwen3-8B starts at 0.1, but increases to 0.4 and 1.0 after one- and two-layer pruning with ShortGPT, respectively. Although SFT reduces the loop fraction to near zero, this improvement is superficial: qualitative analysis (Figure~\ref{fig:example_ft}) reveals that SFT resolves syntactic text loops but leaves underlying logic loops intact. To capture this structural repetition, we utilize Self-BLEU~\citep{self-bleu}, which measures the internal diversity of a corpus by computing the average similarity of each sentence to all others. As detailed in Table~\ref{tab:selfbleu}, pruned models exhibit higher Self-BLEU scores, indicating collapsed output diversity. Importantly, SFT does not restore this diversity. This suggests that SFT fails to relieve repetitive reasoning paths effectively but disguising, rather than repairing, the fundamental reasoning degradation caused by layer pruning.

\begin{figure}[t]
\footnotesize
\centering
\begin{tabular}{@{}p{\linewidth}@{}}
\toprule
\textbf{[AIME24]} \\
Let $p$ be the least prime number for which there exists a positive integer $n$ such that $n^{4}+1$ is divisible by $p^{2}$. Find the least positive integer $m$ such that $m^{4}+1$ is divisible by $p^{2}$. \textbf{Answer: }110 \\

~

\textbf{Before Finetuning (Pruned Only)} \\
\textcolor{teal}{The smallest prime $p$ congruent to 1 modulo 8 is 17... }\textcolor{defaultlightblue}{17 divides 83521-17=83504. Therefore, 17 divides 83521-17, so 17 divides 83521-17. Therefore, 17 divides 83521-17. Therefore, 17 divides 83521-17...} \\

~

\textbf{After Finetuning (Pruned + SFT)} \\
\textcolor{teal}{Let check small primes. If $p=2$, $n^4+1 \equiv 2 \pmod{4}$. If $p=3$, $n^4+1 \equiv 2 \pmod{9}$.}
\textcolor{midblue}{\dots If $p=11$, $n^4 \equiv -1 \equiv 10\ (mod\ 11)$, if $n=5$, $n^4 + 1 = 626\equiv 10\ (\underline{mod\ 11})$\dots The least prime $p$ for which there exists $n$ such that $p^2 \mid n^4+1$ is $p=11$. When $p=11$, we need $n^4 \equiv 10\ (\underline{mod\ 11})$. If $n=5$, $n^4 = 625 \equiv 10\ (\underline{mod\ 11})$.}
\textcolor{defaultblue}{Now we need $m^4+1 \equiv 0\ (mod\ 121)$. If $m=5$, $5^4+1 = 626\ \underline{\equiv 120\ (mod\ 121)}$. Final Answer: \boxed{5}.} \\

\bottomrule
\end{tabular}
\caption{Examples of thinking process from 1-layer pruned s1.1-7B model before and after SFT. \textcolor{defaultlightblue}{Light blue} denotes a verbatim repetition loop where the model repeats an identical string. \textcolor{midblue}{Star command blue} highlights a logical loop where the model repeatedly cycles through the same reasoning steps without progressing. \textcolor{defaultblue}{Blue} indicates a flawed solution path leading to an incorrect result. }
\label{fig:example_ft}
\end{figure}
\begin{table*}[h]
\centering
\scriptsize
\setlength{\tabcolsep}{2.5pt}
\resizebox{\linewidth}{!}{
\begin{tabular}{l|c|cccc|cccc|cccc}
\toprule
& \multicolumn{1}{c|}{\textbf{Original}} 
& \multicolumn{4}{c|}{\textbf{ShortGPT}} 
& \multicolumn{4}{c|}{\textbf{Reverse-order}} 
& \multicolumn{4}{c}{\textbf{LaCo}} \\
\cmidrule(lr){2-2} \cmidrule(lr){3-6} \cmidrule(lr){7-10} \cmidrule(lr){11-14}
\textbf{Model/Method} 
& \textbf{original} 
& \textbf{1 layer} & \textbf{2 layer} & \textbf{1 layer SFT} & \textbf{2 layer SFT} 
& \textbf{1 layer} & \textbf{2 layer} & \textbf{1 layer SFT} & \textbf{2 layer SFT} 
& \textbf{1 layer} & \textbf{2 layer} & \textbf{1 layer SFT} & \textbf{2 layer SFT} \\
\midrule
\textbf{s1.1-7B}  & 0.685 & 0.894 & 0.871 & 0.775 & 0.779 & 0.521 & 0.442 & 0.765 & 0.728 & 0.812 & 0.936 & 0.776 & 0.786 \\
\textbf{Qwen3-8B} & 0.688 & 0.805 & 0.961 & 0.745 & 0.739 & 0.601 & 0.606  & 0.749 & 0.725 & 0.586 & 0.674 & 0.765 & 0.754 \\
\bottomrule

\end{tabular}
}
\caption{Self-BLEU scores for pruned models ($\downarrow$). ``1 layer'' and ``2 layer'' indicate $\#$ of pruned layers, whereas ``SFT'' refer to Full FT.}
\label{tab:selfbleu}
\end{table*}

\subsection{Ablating Sampling Parameters: Temperature, Repetition and Frequency Penalties}

The above observations raise a natural follow-up question: \textit{Could this degradation be alleviated by adjusting sampling strategies at inference time?} We test this aspect by sweeping three common sampling settings: temperature, repetition penalty, and frequency penalty.

\textbf{Settings.} We evaluate sequential test-time scaling on AIME24 with pruned Qwen3-8B.  Results are averaged over  samples by the three pruning methods and all other settings follow our main protocol. (1) \textbf{Temperature}. We sweep temperature $T \in \{0, 0.4, 0.7, 1.0, 1.5, 2.0, 3.0\}$, where higher temperature increases sampling uncertainty. (2) \textbf{Repetition penalty}. We sweep repetition penalty $r \in \{1.0, 1.1, 1.2, 1.5\}$, which penalizes generating tokens that already appeared in the prompt or generated prefix. Values $>1$ discourage repetition. (3) \textbf{Frequency Penalty}. We sweep frequency penalty $\lambda \in \{0, 0.1, 0.3, 0.5, 0.7, 0.9\}$, which penalizes tokens proportionally to how often they have appeared in the generated text so far. Positive values encourage novel tokens.

\textbf{Results.} Results are shown in Tables~\ref{tab:aime24_temp_pruned_qwen3_8b}, \ref{tab:aime24_rep_pruned_qwen3_8b}, \ref{tab:aime24_freq_pruned_qwen3_8b} in Appendix \ref{decoding}. Across all sweeps, we observe a consistent pattern: \textbf{sampling adjustments do not recover pruning-damaged test-time scaling}, and aggressive interventions often worsen performance.
First, temperature exhibits a narrow ``sweet spot'' but no true recovery. Moderate stochasticity improves over greedy decoding: performance peaks around $[0.7, 1.0]$ after pruning. However, higher temperature does not restore reasoning quality and can be catastrophic: at $T\ge 2.0$ both models collapse to near-zero accuracy.
Second, token-level repetition penalties (repetition/frequency) are ineffective and often harmful. For 1-layer pruning, raising the repetition penalty sharply reduces accuracy (e.g., $0.4778 \to 0.1222$ at 8192 tokens as $r: 1.0 \to 1.5$); for 2-layer pruning, $r=1.5$ collapses to $0$ across all budgets. Frequency penalty mirrors this: $\lambda=0$ is best (e.g., $0.4333$ at 8192 tokens for 1-layer), while even $\lambda=0.1$ substantially degrades performance (e.g., $0.1556$), and larger $\lambda$ drives accuracy toward zero.

Overall, sweeping temperature and penalty-based sampling does not restore pruning-induced losses in test-time scaling. Any gains from moderate temperature are limited, while more aggressive stochasticity or repetition penalties often worsen performance, indicating that layer pruning triggers a deeper representational deficit rather than an issue that can be corrected through simple sampling choices.

\begin{table*}[t]
\centering
\scriptsize
\setlength{\tabcolsep}{2.5pt}
\resizebox{\linewidth}{!}{%
\begin{tabular}{l|c|cccc|cccc|cccc}
\toprule
& \multicolumn{1}{c}{\textbf{Original}} 
& \multicolumn{4}{c|}{\textbf{ShortGPT}} 
& \multicolumn{4}{c|}{\textbf{Reverse-order}} 
& \multicolumn{4}{c}{\textbf{LaCo}} \\
\cmidrule(lr){2-2} \cmidrule(lr){3-6} \cmidrule(lr){7-10} \cmidrule(lr){11-14}
\textbf{Behavior/Models} 
& \textbf{Original} 
& \textbf{1 layer} & \textbf{2 layer} & \textbf{1 layer SFT} & \textbf{2 layer SFT} 
& \textbf{1 layer} & \textbf{2 layer} & \textbf{1 layer SFT} & \textbf{2 layer SFT} 
& \textbf{1 layer} & \textbf{2 layer} & \textbf{1 layer SFT} & \textbf{2 layer SFT} \\
\midrule
\textbf{Verification} & 2.167 & 0.867 & 0.467 & 2.333 & 2.033   & 0.433 & 0.000 & 1.967 & 2.000 & 0.867 & 0.900 & 2.033 & 2.700 \\
\textbf{Backtracking} & 2.067 & 0.733 & 0.533 & 2.300 & 2.333 & 1.733 & 0.567 & 1.967 & 2.767 & 1.667 & 1.000 & 1.833 & 2.467 \\
\textbf{Subgoal}      & 4.967 & 2.367 & 1.233 & 4.100 & 3.400 & 4.200 & 2.067 & 3.833 & 2.967 & 3.100 & 0.167 & 4.067 & 3.167 \\
\bottomrule
\end{tabular}%
}

\caption{Self-reflection heuristics of pruned models across different strategies ($\uparrow$). 
``1 layer'' and ``2 layer'' indicate $\#$ of pruned layers,  whereas ``SFT'' refer to Full FT after pruning.}
\label{tab:self_reflection}
\end{table*}


\tikzset{
  seriesBase/.style={semithick, mark=*, mark size=1.6pt},
  seriesK/.style={semithick, dashed, mark=square*, mark size=1.6pt},
}
\pgfplotsset{
  every axis/.append style={
    xmode=log, log basis x=2,
    xtick={512,1024,2048,4096,8192},
    xticklabels={512,1024,2048,4096,8192},
    grid=major, grid style={dashed,gray!30},
    tick style={line width=0.6pt},
    label style={font=\small, fill=none},
    tick label style={font=\small, fill=none},
    title style={font=\bfseries\small, fill=none},
    axis background/.style={fill=none, draw=none},
    ylabel={Accuracy},
    width=0.35\textwidth, height=0.27\textwidth,
    scaled y ticks=false,
    y tick label style={/pgf/number format/fixed,/pgf/number format/precision=2},
  }
}

\def\BaseCoords{(512,0.044) (1024,0.067) (2048,0.133) (4096,0.089) (8192,0.144)}

\def\KZero{(512,0.0) (1024,0.0) (2048,0.0) (4096,0.0) (8192,0.0)}
\def\KOne{(512,0.067) (1024,0.055) (2048,0.078) (4096,0.100) (8192,0.089)}
\def\KTwo{(512,0.089) (1024,0.056) (2048,0.056) (4096,0.044) (8192,0.078)}
\def\KThree{(512,0.089) (1024,0.044) (2048,0.044) (4096,0.056) (8192,0.111)}
\def\KFour{(512,0.089) (1024,0.044) (2048,0.033) (4096,0.111) (8192,0.100)}
\def\KFive{(512,0.044) (1024,0.111) (2048,0.111) (4096,0.111) (8192,0.122)}
\def\KSix{(512,0.100) (1024,0.089) (2048,0.044) (4096,0.111) (8192,0.100)}
\def\KSeven{(512,0.100) (1024,0.111) (2048,0.044) (4096,0.056) (8192,0.111)}
\def\KEight{(512,0.089) (1024,0.089) (2048,0.067) (4096,0.111) (8192,0.111)}
\def\KNine{(512,0.111) (1024,0.111) (2048,0.056) (4096,0.100) (8192,0.067)}
\def\KTen{(512,0.078) (1024,0.111) (2048,0.067) (4096,0.089) (8192,0.078)}
\def\KEleven{(512,0.100) (1024,0.089) (2048,0.111) (4096,0.100) (8192,0.100)}
\def\KTwelve{(512,0.133) (1024,0.089) (2048,0.067) (4096,0.122) (8192,0.122)}
\def\KThirteen{(512,0.033) (1024,0.022) (2048,0.033) (4096,0.044) (8192,0.033)}
\def\KFourteen{(512,0.033) (1024,0.056) (2048,0.033) (4096,0.044) (8192,0.033)}
\def\KFifteen{(512,0.011) (1024,0.011) (2048,0.056) (4096,0.044) (8192,0.044)}
\def\KSixteen{(512,0.022) (1024,0.056) (2048,0.044) (4096,0.033) (8192,0.044)}
\def\KSeventeen{(512,0.022) (1024,0.0) (2048,0.044) (4096,0.033) (8192,0.078)}
\def\KEighteen{(512,0.056) (1024,0.056) (2048,0.022) (4096,0.022) (8192,0.067)}
\def\KNineteen{(512,0.011) (1024,0.033) (2048,0.044) (4096,0.044) (8192,0.044)}
\def\KTwenty{(512,0.011) (1024,0.022) (2048,0.011) (4096,0.011) (8192,0.011)}
\def\KTwentyOne{(512,0.033) (1024,0.033) (2048,0.022) (4096,0.033) (8192,0.044)}
\def\KTwentyTwo{(512,0.0) (1024,0.022) (2048,0.022) (4096,0.044) (8192,0.033)}
\def\KTwentyThree{(512,0.089) (1024,0.111) (2048,0.122) (4096,0.122) (8192,0.156)}
\def\KTwentyFour{(512,0.111) (1024,0.022) (2048,0.056) (4096,0.056) (8192,0.044)}
\def\KTwentyFive{(512,0.044) (1024,0.033) (2048,0.011) (4096,0.033) (8192,0.033)}
\def\KTwentySix{(512,0.100) (1024,0.044) (2048,0.100) (4096,0.067) (8192,0.100)}
\def\KTwentySeven{(512,0.056) (1024,0.078) (2048,0.089) (4096,0.078) (8192,0.078)}

\newcommand{\PlotOne}[2]{%
  \nextgroupplot[title={#1}]
  \addplot+[seriesBase] coordinates \BaseCoords;
  \addplot+[seriesK]    coordinates #2;
}


\begin{figure}[t]
\centering
\begin{tikzpicture}
\begin{groupplot}[
  group style={group size=3 by 3, horizontal sep=1.3cm, vertical sep=1.4cm, ylabels at=edge left},
  ylabel={Accuracy}
]
\nextgroupplot[title={k=19th}, legend to name=legendK, legend columns=2, 
  legend style={draw=none, /tikz/every even column/.style={column sep=0.8em}, font=\small}]

\addplot+[seriesBase] coordinates \BaseCoords;
\addlegendentry{No Pruning}
\addplot+[seriesK] coordinates \KZero;
\addlegendentry{Prune k-th layer}
\PlotOne{k=20th}{\KTwenty}
\PlotOne{k=21st}{\KTwentyOne}
\PlotOne{k=22nd}{\KTwentyTwo}
\PlotOne{k=23rd}{\KTwentyThree}
\PlotOne{k=24th}{\KTwentyFour}
\PlotOne{k=25th}{\KTwentyFive}
\PlotOne{k=26th}{\KTwentySix}
\PlotOne{k=27th}{\KTwentySeven}

\nextgroupplot[hide axis]
\nextgroupplot[hide axis]

\end{groupplot}
\end{tikzpicture}

\vspace{0.4em}
\pgfplotslegendfromname{legendK}
\caption{Brute-Force of s1.1-7B Layer Ablation for Sequential Test-time scaling (last 9 layers)}
\label{fig:brute_force_example}
\end{figure}

\subsection{Assessing Self-Reflection Heuristics} \label{section_5.2}

To further assess reasoning quality, we use a rubric to quantify the frequency of three self-reflection heuristics~\citep{self-reflection} in s1.1-7B on AIME24, with \textit{gpt-4o-mini} serving as an automated judge. The heuristics are: (i) \textbf{verification}, checking the validity of intermediate or final solutions, (ii) \textbf{backtracking}, revising earlier steps or pursuing alternatives after detecting an error, and (iii) \textbf{subgoal setting}, decomposing a complex task into manageable subproblems. These behaviors serve as established proxies for self-reflection, demonstrating a model's ability to monitor, evaluate, and regulate its own reasoning process.

After filtering out non-productive loops and repetitions, we observe a modest yet consistent decline in these heuristics for the pruned model as shown in Table~\ref{tab:self_reflection}. This indicates that pruning undermines more than surface-level diversity: it erodes the model's capacity for structured, self-correcting reasoning. Moreover, this impairment proves resistant to recovery: in most cases, Full FT fails to restore the heuristic frequency to the level of the unpruned baseline, suggesting a more fundamental degradation of capacity for reasoning.

\subsection{Brute-Force Exploration of Layer Ablation} \label{brute-force}

To systematically assess the contribution of all individual layers, we conduct a brute-force ablation study: pruning one layer at a time and evaluating the resulting models on sequential test-time scaling with AIME24. This setup allows us to directly measure the marginal role of each layer in sustaining test-time scaling. Comprehensive results are reported in Appendix~\ref{Appendix_brute_force} (Figures~\ref{fig:brute-force1}, \ref{fig:brute-force2}, \ref{fig:brute-force3}, and \ref{fig:brute-force4}).  Figure~\ref{fig:brute_force_example} highlights the case of s1.1-7B, focusing on ablations of its last nine layers for illustrative purposes. Our findings reveal that almost all the layers play a non-trivial role in enabling test-time scaling. Even the removal of a single layer often leads to substantial degradation, indicating that reasoning relies on widely distributed contributions rather than being concentrated in a small subset of layers. Consequently, even modest layer pruning can disproportionately impair the capacity for test-time scaling, underscoring a fundamental tradeoff between pruning and reasoning fidelity.



\section{Conclusion}

This work provides new insights into how layer pruning reshapes long-chain reasoning in LLMs through the lens of test-time scaling. We find that even minimal layer removal can collapse test-time scaling, especially sequential test-time scaling, causing severe failures on reasoning-intensive benchmarks despite stability on knowledge tasks. Moreover, small-scale supervised fine-tuning has limited effect on recovering the lost test-time scaling. Our in-depth analyses reveal recurring loops, reduced trajectory diversity and diminished self-reflection of model generation after layer pruning, pointing to structural damage rather than surface-level accuracy loss. Overall, our results highlight a fundamental trade-off: efficiency gains from layer pruning often undermine the very mechanisms that enable strong reasoning. We call for caution in applying layer pruning to reasoning-centric settings and advocate for layer pruning methods that explicitly safeguard test-time scaling. Future work could explore hybrid strategies that balance efficiency with robustness, ensuring that pruning preserves both performance and reasoning depth.



\bibliographystyle{colm2026_conference}
\bibliography{colm2026_conference}

@article{shortenedllama,
  title={Shortened llama: A simple depth pruning for large language models},
  author={Kim, Bo-Kyeong and Kim, Geonmin and Kim, Tae-Ho and Castells, Thibault and Choi, Shinkook and Shin, Junho and Song, Hyoung-Kyu},
  journal={arXiv preprint arXiv:2402.02834},
  year={2024}
}

@article{shortgpt,
  title={Shortgpt: Layers in large language models are more redundant than you expect},
  author={Men, Xin and Xu, Mingyu and Zhang, Qingyu and Wang, Bingning and Lin, Hongyu and Lu, Yaojie and Han, Xianpei and Chen, Weipeng},
  journal={ACL Findings},
  year={2025}
}

@article{mka,
  title={Pruning via merging: Compressing llms via manifold alignment based layer merging},
  author={Liu, Deyuan and Qin, Zhanyue and Wang, Hairu and Yang, Zhao and Wang, Zecheng and Rong, Fangying and Liu, Qingbin and Hao, Yanchao and Chen, Xi and Fan, Cunhang and others},
  journal={EMNLP},
  year={2024}
}

@article{laco,
  title={Laco: Large language model pruning via layer collapse},
  author={Yang, Yifei and Cao, Zouying and Zhao, Hai},
  journal={EMNLP Findings},
  year={2024}
}

@article{cot,
  title={Chain-of-thought prompting elicits reasoning in large language models},
  author={Wei, Jason and Wang, Xuezhi and Schuurmans, Dale and Bosma, Maarten and Xia, Fei and Chi, Ed and Le, Quoc V and Zhou, Denny and others},
  journal={Advances in Neural Information Processing Systems},
  year={2022}
}

@article{scalinglaws,
  title={Simple and provable scaling laws for the test-time compute of large language models},
  author={Chen, Yanxi and Pan, Xuchen and Li, Yaliang and Ding, Bolin and Zhou, Jingren},
  journal={arXiv preprint arXiv:2411.19477},
  year={2024}
}

@article{deepseek-r1,
  title={Deepseek-r1: Incentivizing reasoning capability in llms via reinforcement learning},
  author={Guo, Daya and Yang, Dejian and Zhang, Haowei and Song, Junxiao and Zhang, Ruoyu and Xu, Runxin and Zhu, Qihao and Ma, Shirong and Wang, Peiyi and Bi, Xiao and others},
  journal={arXiv preprint arXiv:2501.12948},
  year={2025}
}

@article{demystifying,
  title={Demystifying long chain-of-thought reasoning in llms},
  author={Yeo, Edward and Tong, Yuxuan and Niu, Morry and Neubig, Graham and Yue, Xiang},
  journal={International Conference on Learning Representations},
  year={2025}
}

@article{kimi,
  title={Kimi k1. 5: Scaling reinforcement learning with llms},
  author={Team, Kimi and Du, Angang and Gao, Bofei and Xing, Bowei and Jiang, Changjiu and Chen, Cheng and Li, Cheng and Xiao, Chenjun and Du, Chenzhuang and Liao, Chonghua and others},
  journal={arXiv preprint arXiv:2501.12599},
  year={2025}
}

@article{internvl,
  title={Expanding performance boundaries of open-source multimodal models with model, data, and test-time scaling},
  author={Chen, Zhe and Wang, Weiyun and Cao, Yue and Liu, Yangzhou and Gao, Zhangwei and Cui, Erfei and Zhu, Jinguo and Ye, Shenglong and Tian, Hao and Liu, Zhaoyang and others},
  journal={arXiv preprint arXiv:2412.05271},
  year={2024}
}

@article{s1,
  title={s1: Simple test-time scaling},
  author={Muennighoff, Niklas and Yang, Zitong and Shi, Weijia and Li, Xiang Lisa and Fei-Fei, Li and Hajishirzi, Hannaneh and Zettlemoyer, Luke and Liang, Percy and Cand\`{e}s, Emmanuel and Hashimoto, Tatsunori},
  journal={arXiv preprint arXiv:2501.19393},
  year={2025}
}

@article{t1,
  title={Advancing language model reasoning through reinforcement learning and inference scaling},
  author={Hou, Zhenyu and Lv, Xin and Lu, Rui and Zhang, Jiajie and Li, Yujiang and Yao, Zijun and Li, Juanzi and Tang, Jie and Dong, Yuxiao},
  journal={arXiv preprint arXiv:2501.11651},
  year={2025}
}

@article{ttrl,
  title={Ttrl: Test-time reinforcement learning},
  author={Zuo, Yuxin and Zhang, Kaiyan and Sheng, Li and Qu, Shang and Cui, Ganqu and Zhu, Xuekai and Li, Haozhan and Zhang, Yuchen and Long, Xinwei and Hua, Ermo and others},
  journal={arXiv preprint arXiv:2504.16084},
  year={2025}
}

@article{ps1,
  title={Universal self-consistency for large language model generation},
  author={Chen, Xinyun and Aksitov, Renat and Alon, Uri and Ren, Jie and Xiao, Kefan and Yin, Pengcheng and Prakash, Sushant and Sutton, Charles and Wang, Xuezhi and Zhou, Denny},
  journal={arXiv preprint arXiv:2311.17311},
  year={2023}
}

@article{ps2,
  title={Monte carlo tree search boosts reasoning via iterative preference learning},
  author={Xie, Yuxi and Goyal, Anirudh and Zheng, Wenyue and Kan, Min-Yen and Lillicrap, Timothy P and Kawaguchi, Kenji and Shieh, Michael},
  journal={arXiv preprint arXiv:2405.00451},
  year={2024}
}

@inproceedings{ps4,
  title={Openprm: Building open-domain process-based reward models with preference trees},
  author={Zhang, Kaiyan and Zhang, Jiayuan and Li, Haoxin and Zhu, Xuekai and Hua, Ermo and Lv, Xingtai and Ding, Ning and Qi, Biqing and Zhou, Bowen},
  booktitle={International Conference on Learning Representations},
  year={2025}
}

@article{reassessing,
  title={Reassessing layer pruning in llms: New insights and methods},
  author={Lu, Yao and Cheng, Hao and Fang, Yujie and Wang, Zeyu and Wei, Jiaheng and Xu, Dongwei and Xuan, Qi and Yang, Xiaoniu and Zhu, Zhaowei},
  journal={International Conference on Learning Representations},
  year={2026}
}

@article{qwen3,
  title={Qwen3 technical report},
  author={Yang, An and Li, Anfeng and Yang, Baosong and Zhang, Beichen and Hui, Binyuan and Zheng, Bo and Yu, Bowen and Gao, Chang and Huang, Chengen and Lv, Chenxu and others},
  journal={arXiv preprint arXiv:2505.09388},
  year={2025}
}

@article{qwen2.5,
  title={Qwen2. 5-math technical report: Toward mathematical expert model via self-improvement},
  author={Yang, An and Zhang, Beichen and Hui, Binyuan and Gao, Bofei and Yu, Bowen and Li, Chengpeng and Liu, Dayiheng and Tu, Jianhong and Zhou, Jingren and Lin, Junyang and others},
  journal={arXiv preprint arXiv:2409.12122},
  year={2024}
}

@inproceedings{gpqa,
  title={Gpqa: A graduate-level google-proof q\&a benchmark},
  author={Rein, David and Hou, Betty Li and Stickland, Asa Cooper and Petty, Jackson and Pang, Richard Yuanzhe and Dirani, Julien and Michael, Julian and Bowman, Samuel R},
  booktitle={COLM},
  year={2024}
}

@inproceedings{math500,
  title={Let's verify step by step},
  author={Lightman, Hunter and Kosaraju, Vineet and Burda, Yuri and Edwards, Harrison and Baker, Bowen and Lee, Teddy and Leike, Jan and Schulman, John and Sutskever, Ilya and Cobbe, Karl},
  booktitle={International Conference on Learning Representations},
  year={2023}
}

@misc{AIME24,
  title        = {American Invitational Mathematics Examination 2024 AIME},
  howpublished = {\url{https://artofproblemsolving.com/wiki/index.php/AIME_Problems_and_Solutions}},
  year         = {2024},
  author       = {{Mathematical Association of America}}
}

@misc{harness,
  author       = {Gao, Leo and Tow, Jonathan and Abbasi, Baber and Biderman, Stella and Black, Sid and DiPofi, Anthony and Foster, Charles and Golding, Laurence and Hsu, Jeffrey and Le Noac'h, Alain and Li, Haonan and McDonell, Kyle and Muennighoff, Niklas and Ociepa, Chris and Phang, Jason and Reynolds, Laria and Schoelkopf, Hailey and Skowron, Aviya and Sutawika, Lintang and Tang, Eric and Thite, Anish and Wang, Ben and Wang, Kevin and Zou, Andy},
  title        = {The Language Model Evaluation Harness},
  month        = 07,
  year         = 2024,
  publisher    = {Zenodo},
  version      = {v0.4.3},
  doi          = {10.5281/zenodo.12608602},
  url          = {https://zenodo.org/records/12608602}
}

@article{lora,
  title={Lora: Low-rank adaptation of large language models.},
  author={Hu, Edward J and Shen, Yelong and Wallis, Phillip and Allen-Zhu, Zeyuan and Li, Yuanzhi and Wang, Shean and Wang, Lu and Chen, Weizhu and others},
  journal={International Conference on Learning Representations},
  year={2022}
}

@article{mmlu,
  title={Measuring massive multitask language understanding},
  author={Hendrycks, Dan and Burns, Collin and Basart, Steven and Zou, Andy and Mazeika, Mantas and Song, Dawn and Steinhardt, Jacob},
  journal={International Conference on Learning Representations},
  year={2021}
}

@inproceedings{zhang2019improving,
  title={Improving deep transformer with depth-scaled initialization and merged attention},
  author={Zhang, Biao and Titov, Ivan and Sennrich, Rico},
  booktitle={Proceedings of the 2019 conference on empirical methods in natural language processing and the 9th international joint conference on natural language processing (EMNLP-IJCNLP)},
  pages={898--909},
  year={2019}
}

@inproceedings{song2026demystifying,
  title={Demystifying the roles of llm layers in retrieval, knowledge, and reasoning},
  author={Song, Xinyuan and Wang, Keyu and Li, PengXiang and Yin, Lu and Liu, Shiwei},
  booktitle={ICASSP 2026-2026 IEEE International Conference on Acoustics, Speech and Signal Processing (ICASSP)},
  pages={16792--16796},
  year={2026},
  organization={IEEE}
}

@article{dey2026don,
  title={Don't be lazy: CompleteP enables compute-efficient deep transformers},
  author={Dey, Nolan and Zhang, Bin and Noci, Lorenzo and Li, Mufan and Bordelon, Blake and Bergsma, Shane and Pehlevan, Cengiz and Hanin, Boris and Hestness, Joel},
  journal={Advances in Neural Information Processing Systems},
  volume={38},
  pages={137707--137739},
  year={2026}
}

@article{takase2023spike,
  title={Spike no more: Stabilizing the pre-training of large language models},
  author={Takase, Sho and Kiyono, Shun and Kobayashi, Sosuke and Suzuki, Jun},
  journal={arXiv preprint arXiv:2312.16903},
  year={2023}
}

@article{lns,
  title={The curse of depth in large language models},
  author={Sun, Wenfang and Song, Xinyuan and Li, Pengxiang and Yin, Lu and Zheng, Yefeng and Liu, Shiwei},
  journal={arXiv preprint arXiv:2502.05795},
  year={2025}
}

@article{gptailor,
  title={GPTailor: Large Language Model Pruning Through Layer Cutting and Stitching},
  author={Su, Guinan and Shen, Li and Yin, Lu and Liu, Shiwei and Yang, Yanwu and Geiping, Jonas},
  journal={arXiv preprint arXiv:2506.20480},
  year={2025}
}

@article{compact,
  title={Compact language models via pruning and knowledge distillation},
  author={Muralidharan, Saurav and Turuvekere Sreenivas, Sharath and Joshi, Raviraj and Chochowski, Marcin and Patwary, Mostofa and Shoeybi, Mohammad and Catanzaro, Bryan and Kautz, Jan and Molchanov, Pavlo},
  journal={Advances in Neural Information Processing Systems},
  year={2024}
}

@article{chen2026post,
  title={Post-LayerNorm Is Back: Stable, ExpressivE, and Deep},
  author={Chen, Chen and Wei, Lai},
  journal={arXiv preprint arXiv:2601.19895},
  year={2026}
}

@article{muhtar2026does,
  title={When Does Sparsity Mitigate the Curse of Depth in LLMs},
  author={Muhtar, Dilxat and Song, Xinyuan and Pokutta, Sebastian and Zimmer, Max and Pelleriti, Nico and Hofmann, Thomas and Liu, Shiwei},
  journal={arXiv preprint arXiv:2603.15389},
  year={2026}
}

@inproceedings{trochelmann2026kitenorm,
  title={KiteNorm: Variance Regularisation for Stable and Scalable Post-LN Transformers},
  author={Trochelmann, Leon A and Movahedi, Sajad and Liu, Shiwei and Orvieto, Antonio},
  booktitle={High-dimensional Learning Dynamics 2026}
}

@article{li2024mix,
  title={Mix-ln: Unleashing the power of deeper layers by combining pre-ln and post-ln},
  author={Li, Pengxiang and Yin, Lu and Liu, Shiwei},
  journal={arXiv preprint arXiv:2412.13795},
  year={2024}
}

@article{bbh,
  title={Challenging BIG-Bench Tasks and Whether Chain-of-Thought Can Solve Them},
  author={Suzgun, Mirac and Scales, Nathan and Sch\"{a}rli, Nathanael and Gehrmann, Sebastian and Tay, Yi and Chung, Hyung Won and Chowdhery, Aakanksha and Le, Quoc V and Chi, Ed H and Zhou, Denny and and Wei, Jason},
  journal={ACL Findings},
  year={2023}
}

@article{gsm8k,
  title={Training verifiers to solve math word problems},
  author={Cobbe, Karl and Kosaraju, Vineet and Bavarian, Mohammad and Chen, Mark and Jun, Heewoo and Kaiser, Lukasz and Plappert, Matthias and Tworek, Jerry and Hilton, Jacob and Nakano, Reiichiro and others},
  journal={arXiv preprint arXiv:2110.14168},
  year={2021}
}

@inproceedings{self-bleu,
  title={Texygen: A benchmarking platform for text generation models},
  author={Zhu, Yaoming and Lu, Sidi and Zheng, Lei and Guo, Jiaxian and Zhang, Weinan and Wang, Jun and Yu, Yong},
  booktitle={The 41st international ACM SIGIR conference on research \& development in information retrieval},
  pages={1097--1100},
  year={2018}
}

@article{llm-pruner,
  title={Llm-pruner: On the structural pruning of large language models},
  author={Ma, Xinyin and Fang, Gongfan and Wang, Xinchao},
  journal={Advances in Neural Information Processing Systems},
  year={2023}
}

@article{depth,
  title={A deeper look at depth pruning of llms},
  author={Siddiqui, Shoaib Ahmed and Dong, Xin and Heinrich, Greg and Breuel, Thomas and Kautz, Jan and Krueger, David and Molchanov, Pavlo},
  journal={arXiv preprint arXiv:2407.16286},
  year={2024}
}

@article{self-reflection,
  title={Cognitive behaviors that enable self-improving reasoners, or, four habits of highly effective stars},
  author={Gandhi, Kanishk and Chakravarthy, Ayush and Singh, Anikait and Lile, Nathan and Goodman, Noah D},
  journal={arXiv preprint arXiv:2503.01307},
  year={2025}
}

@article{slicegpt,
  title={Slicegpt: Compress large language models by deleting rows and columns},
  author={Ashkboos, Saleh and Croci, Maximilian L and Nascimento, Marcelo Gennari do and Hoefler, Torsten and Hensman, James},
  journal={arXiv preprint arXiv:2401.15024},
  year={2024}
}

@article{finercut,
  title={Finercut: Finer-grained interpretable layer pruning for large language models},
  author={Zhang, Yang and Li, Yawei and Wang, Xinpeng and Shen, Qianli and Plank, Barbara and Bischl, Bernd and Rezaei, Mina and Kawaguchi, Kenji},
  journal={arXiv preprint arXiv:2405.18218},
  year={2024}
}

@article{sleb,
  title={Sleb: Streamlining llms through redundancy verification and elimination of transformer blocks},
  author={Song, Jiwon and Oh, Kyungseok and Kim, Taesu and Kim, Hyungjun and Kim, Yulhwa and Kim, Jae-Joon},
  journal={International Conference on Machine Learning},
  year={2024}
}

@article{transformer,
  title={Attention is all you need},
  author={Vaswani, Ashish and Shazeer, Noam and Parmar, Niki and Uszkoreit, Jakob and Jones, Llion and Gomez, Aidan N and Kaiser, {\L}ukasz and Polosukhin, Illia},
  journal={Advances in neural information processing systems},
  volume={30},
  year={2017}
}

@misc{loopfraction,
      title={Wait, Wait, Wait... Why Do Reasoning Models Loop?}, 
      author={Charilaos Pipis and Shivam Garg and Vasilis Kontonis and Vaishnavi Shrivastava and Akshay Krishnamurthy and Dimitris Papailiopoulos},
      year={2025},
      eprint={2512.12895},
      archivePrefix={arXiv},
      primaryClass={cs.LG},
      url={https://arxiv.org/abs/2512.12895}, 
}

@inproceedings{blockpruner,
  title={Blockpruner: Fine-grained pruning for large language models},
  author={Zhong, Longguang and Wan, Fanqi and Chen, Ruijun and Quan, Xiaojun and Li, Liangzhi},
  booktitle={Findings of the Association for Computational Linguistics: ACL 2025},
  pages={5065--5080},
  year={2025}
}

@article{tts,
  title={Scaling llm test-time compute optimally can be more effective than scaling model parameters},
  author={Snell, Charlie and Lee, Jaehoon and Xu, Kelvin and Kumar, Aviral},
  journal={arXiv preprint arXiv:2408.03314},
  year={2024}
}

@article{comanici2025gemini,
  title={Gemini 2.5: Pushing the frontier with advanced reasoning, multimodality, long context, and next generation agentic capabilities},
  author={Comanici, Gheorghe and Bieber, Eric and Schaekermann, Mike and Pasupat, Ice and Sachdeva, Noveen and Dhillon, Inderjit and Blistein, Marcel and Ram, Ori and Zhang, Dan and Rosen, Evan and others},
  journal={arXiv preprint arXiv:2507.06261},
  year={2025}
}

@article{selfdistill,
  title={Self-data distillation for recovering quality in pruned large language models},
  author={Thangarasa, Vithursan and Venkatesh, Ganesh and Lasby, Mike and Sinnadurai, Nish and Lie, Sean},
  journal={Proceedings of Machine Learning and Systems},
  volume={7},
  year={2025}
}

\newpage
\appendix

\section{Details about ShortGPT, Reverse-order, LaCo, and Their Implementations} \label{Appendix_pruning}

We select three representative layer pruning strategies, ShortGPT, Reverse-order, and LaCo, due to their superior empirical performance.

\textbf{ShortGPT.} ShortGPT proposes a layer pruning strategy using Block Influence (BI) to indicate layer importance. The BI score of the $i$-th layer is defined as
\[
\text{BI}_i = 1 - \mathbb{E}_{X, t} \frac{X_{i,t}^\top X_{i+1,t}}{\left\| X_{i,t} \right\|_2 \, \left\| X_{i+1,t} \right\|_2},
\]
where $X_i$ denotes the input to the $i$-th layer, and $X_{i,t}$ is the $t$-th row of $X_i$. BI measures the degree of alignment between consecutive layer activations: a higher BI indicates lower inter-layer correlation, suggesting that the corresponding block contributes less to information propagation and can be pruned with limited performance degradation.

\textbf{Reverse-order.} Reverse-order pruning assigns lower importance scores to deeper layers, assuming that tail layers contribute less to the overall performance of LLMs. Because the final output layer often plays a critical role in generation and prediction and pruning it causes a sharp performance drop, in this work, we retain the final layer and begin pruning backward from the penultimate layer.

\textbf{LaCo.} LaCo is a layer-collapsing algorithm merges parameters that are highly similar, thereby reducing redundancy across depth. Specifically, given a target layer $l$ and $m$ subsequent layers to be merged, the merged parameter $\theta_l^{*}$ is computed as
\[
\theta_l^{*} = \theta_l + (\theta_{l+1} - \theta_l) + \cdots + (\theta_{l+m} - \theta_l)
= \theta_l + \sum_{k=1}^{m} (\theta_{l+k} - \theta_l).
\]
During training-free merging, the LaCo algorithm scans all pairs of consecutive layers and evaluates their similarity. Layers whose similarity exceeds a predefined threshold are collapsed, allowing us to control the number of merged layers by adjusting this threshold. We apply this method using both the default dataset from the LaCo study and a dataset sampled from the MATH500 training set.

We also provide the specific layer pruning order used for the pruning and merging experiments throughout this paper, as detailed in Table~\ref{tab:layer_sequence}. To generate these sequences, we followed the calibration methodologies of the original publications. For the ShortGPT method, we used a sample of 10,000 instances from the PG19 dataset. For the LaCo method, we used 5 data samples from wikitext, consistent with its official repository.

\begin{table}[h]
\centering
\begin{tabular}{@{}lccc@{}}
\toprule
\textbf{Model} & \textbf{ShortGPT} & \textbf{Reverse-order} & \textbf{LaCo} \\ \midrule
\textbf{Qwen3-8B} & [20, 17, 21] & [34, 33, 32] & [34, 33, 29] \\ \midrule
\textbf{s1.1-7B} & [16, 17] & [26, 25] & [22, 18] \\ \bottomrule
\end{tabular}
\caption{Layer pruning order used for pruning and merging experiments.}
\label{tab:layer_sequence}
\end{table}

\section{Impact of Calibration Data}
In this section, we discuss the sensitivity of ShortGPT and LaCo to calibration data. For ShortGPT, calibration data has a significant effect. As shown in Table~\ref{tab:shortgpt_cal_data}, reasoning-focused datasets MATH500 and AIME24 tend to lead ShortGPT to prune shallow layers first (e.g., 1--5), while general-purpose text PG19 shifts pruning mainly to middle or deeper layers (14--23). In contrast, the general-purpose LlaMA3.1-8B yields a stable pruning order regardless of calibration data. Our findings indicate that calibration data can affect identifying layer importance notably in specialized models and tasks. Concurrently, LaCo proves more robust. Table~\ref{tab:laco_cal_data} shows that LaCo selects nearly identical sets of layers across domains for LlaMA3.1-8B and Qwen3-8B, with only modest variation in s1.1-7B. Unlike ShortGPT, it avoids drastic variations in identifying layer importance, underscoring that direct-removal pruning method may demand in-domain calibration data, while merging-based pruning method may be comparatively insensitive to calibration choice.

\begin{table*}[h]
\centering
\resizebox{0.8\linewidth}{!}{%
\begin{tabular}{@{}lll@{}}
\toprule
\textbf{Model} & \textbf{Calibration Dataset} & \textbf{Pruning Order (Low Layer Importance First)} \\ \midrule
\textbf{Qwen3-8B} & PG19 & {[}20, 17, 21, 18, 2, 16, 19, 15, 23, 14{]} \\
 & MATH500 & {[}2, 3, 1, 5, 17, 4, 20, 16, 19, 21{]} \\
 & AIME24 & {[}2, 3, 1, 20, 17, 5, 16, 21, 24, 19{]} \\ \midrule
\textbf{s1.1-7B} & PG19 & {[}16, 17, 15, 14, 18, 13, 12, 11, 20, 23{]} \\
 & MATH500 & {[}2, 16, 15, 5, 17, 14, 13, 12, 4, 1{]} \\
 & AIME24 & {[}16, 2, 15, 17, 14, 5, 13, 4, 12, 1{]} \\ \midrule
\textbf{LlaMA3.1-8B} & PG19 & {[}25, 27, 26, 24, 28, 23, 22, 29, 20, 21{]} \\
 & MATH500 & {[}25, 24, 26, 23, 27, 28, 22, 21, 29, 20{]} \\
 & AIME24 & {[}25, 23, 24, 26, 27, 22, 28, 21, 20, 19{]} \\ \bottomrule
\end{tabular}}
\caption{Effect of calibration data on ShortGPT's pruning order. Qwen3-8B and s1.1-7B show high sensitivity to calibration data, while the general-purpose LLaMA3.1-8B is stable.}
\label{tab:shortgpt_cal_data}
\end{table*}

\begin{table*}[h]
\centering
\resizebox{0.8\linewidth}{!}{%
\begin{tabular}{@{}llccc@{}}
\toprule
\textbf{Model} & \textbf{Calibration Dataset} & \textbf{Merge 1 Layer} & \textbf{Merge 2 Layers} & \textbf{Merge 3 Layers} \\ \midrule
\textbf{Qwen3-8B} & Wikitext-2 & {[}34{]} & {[}34, 29{]} & {[}34, 33, 29{]} \\
& MATH500 & {[}34{]} & {[}34, 33{]} & {[}34, 33, 32{]} \\ \midrule
\textbf{s1.1-7B} & Wikitext-2 & {[}22{]} & {[}22, 18{]} & {[}25, 18, 17{]} \\
& MATH500 & {[}21{]} & {[}26, 18{]} & {[}26, 25, 16{]} \\ \midrule
\textbf{LlaMA3.1-8B} & Wikitext-2 & {[}30{]} & {[}30, 15{]} & {[}30, 29, 15{]} \\
& MATH500 & {[}30{]} & {[}30, 27{]} & {[}30, 29, 26{]} \\ \bottomrule
\end{tabular}}
\caption{
    Effect of calibration data on LaCo's merge order. The table shows the sequence of layers to be absorbed. The results show that LaCo's layer selection is notably more stable against changes in calibration data compared to ShortGPT.
}
\label{tab:laco_cal_data}
\end{table*}

\section{Additional Results for Sequential Scaling of s1.1-7B and Its Pruned Varients in Section \ref{section3}} \label{Appendix_scaling}

As shown in Fig \ref{fig-s1-scaling}, in line with Qwen3-8B, s1.1-7B is even more fragile: across ShortGPT, Reverse-order, and LaCo, removing just one layer already noticeably weakens scaling. After pruning two layers, the scaling behavior essentially disappears, performance plateaus, and can even decline as thinking tokens increase.

\begin{figure*}[h]
\centering
\begin{tikzpicture}
\begin{groupplot}[
  group style={group size=3 by 3, horizontal sep=1.2cm, vertical sep=0.7cm},
  width=0.32\textwidth, height=0.24\textwidth,
  xmode=log, log basis x=2,
  xtick={512,1024,2048,4096,8192},
  xticklabels={512,1024,2048,4096,8192},
  grid=major, grid style={dashed,gray!30},
  tick style={line width=0.6pt},
  label style={font=\small, fill=none},
  tick label style={font=\small, fill=none},
  xticklabel style={font=\scriptsize},
  title style={font=\bfseries\small, fill=none},
  axis background/.style={fill=none, draw=none},
  ylabel={Accuracy},
]

\nextgroupplot[
  title={MATH500},
  ylabel={Acc (ShortGPT)},
  legend to name=legendSOneScalingShortGPT,
  legend columns=3,
  legend style={draw=none, /tikz/every even column/.style={column sep=0.8em}, font=\small}
]
\addplot+[seriesA] coordinates {(512,0.626) (1024,0.701) (2048,0.760) (4096,0.783) (8192,0.803)};
\addlegendentry{No pruning}
\addplot+[seriesB] coordinates {(512,0.559) (1024,0.575) (2048,0.617) (4096,0.623) (8192,0.653)};
\addlegendentry{Prune 1 layer}
\addplot+[seriesC] coordinates {(512,0.386) (1024,0.417) (2048,0.428) (4096,0.440) (8192,0.435)};
\addlegendentry{Prune 2 layers}

\nextgroupplot[title={GPQA Diamond}, ylabel={}]
\addplot+[seriesA] coordinates {(512,0.390) (1024,0.353) (2048,0.401) (4096,0.411) (8192,0.428)};
\addplot+[seriesB] coordinates {(512,0.328) (1024,0.318) (2048,0.343) (4096,0.343) (8192,0.365)};
\addplot+[seriesC] coordinates {(512,0.318) (1024,0.300) (2048,0.306) (4096,0.288) (8192,0.311)};

\nextgroupplot[
  title={AIME24}, ylabel={},
  scaled y ticks=false,
  y tick label style={/pgf/number format/fixed,/pgf/number format/precision=2}
]
\addplot+[seriesA] coordinates {(512,0.044) (1024,0.067) (2048,0.133) (4096,0.089) (8192,0.144)};
\addplot+[seriesB] coordinates {(512,0.033) (1024,0.011) (2048,0.044) (4096,0.044) (8192,0.067)};
\addplot+[seriesC] coordinates {(512,0.022) (1024,0.000) (2048,0.000) (4096,0.011) (8192,0.022)};

\nextgroupplot[
  ylabel={Acc (Reverse)},
  legend to name=legendSOneScalingReverse,
  legend columns=3,
  legend style={draw=none, /tikz/every even column/.style={column sep=0.8em}, font=\small}
]
\addplot+[seriesA] coordinates {(512,0.626) (1024,0.701) (2048,0.760) (4096,0.783) (8192,0.803)};
\addlegendentry{No pruning}
\addplot+[seriesB] coordinates {(512,0.472) (1024,0.504) (2048,0.548) (4096,0.602) (8192,0.606)};
\addlegendentry{Prune 1 layer}
\addplot+[seriesC] coordinates {(512,0.136) (1024,0.156) (2048,0.164) (4096,0.169) (8192,0.204)};
\addlegendentry{Prune 2 layers}

\nextgroupplot[ylabel={}]
\addplot+[seriesA] coordinates {(512,0.390) (1024,0.353) (2048,0.401) (4096,0.411) (8192,0.428)};
\addplot+[seriesB] coordinates {(512,0.339) (1024,0.330) (2048,0.364) (4096,0.355) (8192,0.337)};
\addplot+[seriesC] coordinates {(512,0.306) (1024,0.310) (2048,0.293) (4096,0.305) (8192,0.304)};

\nextgroupplot[
  ylabel={}, scaled y ticks=false,
  y tick label style={/pgf/number format/fixed,/pgf/number format/precision=2}
]
\addplot+[seriesA] coordinates {(512,0.044) (1024,0.067) (2048,0.133) (4096,0.089) (8192,0.144)};
\addplot+[seriesB] coordinates {(512,0.033) (1024,0.011) (2048,0.033) (4096,0.056) (8192,0.067)};
\addplot+[seriesC] coordinates {(512,0.011) (1024,0.011) (2048,0.011) (4096,0.022) (8192,0.033)};

\nextgroupplot[
  ylabel={Acc (LaCo)},
  xlabel={Thinking tokens},   
  legend to name=legendSOneScalingLaCo,
  legend columns=3,
  legend style={draw=none, /tikz/every even column/.style={column sep=0.8em}, font=\small}
]
\addplot+[seriesA] coordinates {(512,0.626) (1024,0.701) (2048,0.760) (4096,0.783) (8192,0.803)};
\addlegendentry{No pruning}
\addplot+[seriesB] coordinates {(512,0.234) (1024,0.245) (2048,0.275) (4096,0.270) (8192,0.273)};
\addlegendentry{Prune 1 layer}
\addplot+[seriesC] coordinates {(512,0.143) (1024,0.137) (2048,0.147) (4096,0.138) (8192,0.136)};
\addlegendentry{Prune 2 layers}

\nextgroupplot[ylabel={}, xlabel={Thinking tokens}]
\addplot+[seriesA] coordinates {(512,0.390) (1024,0.353) (2048,0.401) (4096,0.411) (8192,0.428)};
\addplot+[seriesB] coordinates {(512,0.338) (1024,0.321) (2048,0.330) (4096,0.333) (8192,0.343)};
\addplot+[seriesC] coordinates {(512,0.325) (1024,0.273) (2048,0.269) (4096,0.285) (8192,0.289)};

\nextgroupplot[
  ylabel={}, xlabel={Thinking tokens},
  scaled y ticks=false,
  y tick label style={/pgf/number format/fixed,/pgf/number format/precision=2}
]
\addplot+[seriesA] coordinates {(512,0.044) (1024,0.067) (2048,0.133) (4096,0.089) (8192,0.144)};
\addplot+[seriesB] coordinates {(512,0.022) (1024,0.011) (2048,0.033) (4096,0.022) (8192,0.022)};
\addplot+[seriesC] coordinates {(512,0.000) (1024,0.011) (2048,0.044) (4096,0.011) (8192,0.033)};

\end{groupplot}
\end{tikzpicture}

\vspace{0.4em}
\pgfplotslegendfromname{legendSOneScalingShortGPT}

\caption{Sequential test-time scaling of s1.1-7B under different pruning depths. }
\label{fig-s1-scaling}
\end{figure*}

\section{Learning Rate Search in LoRA Fine-tuning} \label{Appendix_lr}

Table \ref{tab:search_lr} presents the results of grid search over learning rates of LoRA fine-tuning. We fine-tune the pruned variants of s1.1-7B and Qwen3-8B on the s1.1-1K dataset \citep{s1} using LoRA with rank 16 and scaling factor 32.
For each pruned configuration, we conduct a grid search over learning rates
$[1\times10^{-6},\, 4\times10^{-6},\, 1\times10^{-5},\, 4\times10^{-5},\, 1\times10^{-4}]$
and select the best-performing value based on validation accuracy on MATH500, GPQA Diamond, and AIME24, respectively. The final results are reported on the corresponding evaluation datasets.

\begin{table}[h]
\centering
\scriptsize
\setlength{\tabcolsep}{2.5pt}
\resizebox{\textwidth}{!}{
\begin{tabular}{l|ccccc|ccccc|ccccc}
\toprule
& \multicolumn{5}{c|}{MATH500} & \multicolumn{5}{c|}{GPQA Diamond} & \multicolumn{5}{c}{AIME24} \\
\cmidrule(lr){2-6} \cmidrule(lr){7-11} \cmidrule(lr){12-16}
Model/Method & 1e-6 & 4e-6 & 1e-5 & 4e-5 & 1e-4 & 1e-6 & 4e-6 & 1e-5 & 4e-5 & 1e-4 & 1e-6 & 4e-6 & 1e-5 & 4e-5 & 1e-4 \\
\midrule
s1.1-7B ShortGPT\_1\_layer   & 0.478 & 0.410 & 0.510 & 0.578 & \textbf{0.608} & 0.273 & 0.267 & 0.252 & 0.313 & \textbf{0.338} & 0.033 & 0.033 & 0.067 & \textbf{0.100} & 0.033 \\
s1.1-7B Reverse\_1\_layer   & 0.377 & 0.403 & 0.382 & \textbf{0.446} & 0.400 & \textbf{0.292} & 0.257 & 0.247 & 0.242 & 0.247 & 0.000 & 0.033 & \textbf{0.033} & 0.033 & 0.033 \\
s1.1-7B LaCo\_1\_layer      & 0.268 & 0.324 & 0.434 & \textbf{0.532} & 0.590 & 0.303 & 0.298 & \textbf{0.303} & 0.303 & 0.293 & 0.033 & 0.033 & 0.066 & \textbf{0.100} & 0.033 \\
s1.1-7B ShortGPT\_2\_layer  & 0.282 & 0.374 & 0.360 & 0.454 & \textbf{0.512} & 0.253 & 0.268 & \textbf{0.308} & 0.262 & 0.293 & 0.000 & 0.000 & 0.000 & \textbf{0.033} & 0.033 \\
s1.1-7B Reverse\_2\_layer   & 0.098 & 0.073 & 0.082 & \textbf{0.132} & 0.104 & 0.247 & 0.249 & 0.268 & \textbf{0.273} & 0.257 & 0.000 & 0.000 & \textbf{0.033} & 0.000 & 0.000 \\
s1.1-7B LaCo\_2\_layer      & 0.198 & 0.192 & 0.302 & 0.386 & \textbf{0.454} & 0.222 & 0.273 & \textbf{0.298} & 0.283 & 0.268 & 0.067 & \textbf{0.067} & 0.033 & 0.033 & 0.000 \\
Qwen3-8B ShortGPT\_1\_layer & 0.876 & \textbf{0.880} & 0.800 & 0.758 & 0.760 & 0.444 & \textbf{0.444} & 0.399 & 0.399 & 0.444 & \textbf{0.433} & 0.400 & 0.300 & 0.133 & 0.167 \\
Qwen3-8B Reverse\_1\_layer  & 0.886 & 0.918 & \textbf{0.926} & 0.822 & 0.828 & 0.477 & \textbf{0.505} & 0.419 & 0.404 & 0.449 & 0.567 & \textbf{0.600} & 0.567 & 0.233 & 0.266 \\
Qwen3-8B LaCo\_1\_layer     & \textbf{0.934} & 0.930 & 0.884 & 0.081 & 0.844 & 0.423 & 0.485 & 0.414 & 0.429 & \textbf{0.510} & 0.167 & 0.167 & 0.133 & \textbf{0.267} & 0.200 \\
Qwen3-8B ShortGPT\_2\_layer & 0.282 & 0.374 & 0.360 & 0.454 & \textbf{0.512} & 0.338 & 0.328 & 0.348 & \textbf{0.374} & 0.343 & \textbf{0.200} & 0.167 & 0.100 & 0.067 & 0.100 \\
Qwen3-8B Reverse\_2\_layer  & 0.808 & 0.918 & \textbf{0.934} & 0.776 & 0.804 & 0.409 & 0.455 & \textbf{0.484} & 0.399 & 0.414 & 0.367 & 0.267 & \textbf{0.567} & 0.467 & 0.200 \\
Qwen3-8B LaCo\_2\_layer     & 0.686 & 0.708 & 0.572 & \textbf{0.718} & 0.710 & 0.342 & 0.378 & 0.343 & 0.369 & \textbf{0.404} & 0.067 & 0.067 & 0.100 & 0.067 & \textbf{0.167} \\
\bottomrule
\end{tabular}
}
\caption{Results of LoRA FT across different learning rates on MATH500, GPQA Diamond, and AIME24. Best results per row are in \textbf{bold}. ``xxx\_k\_layer'' means ``pruned k layers by xxx'' and ``Reverse'' denotes Reverse-order pruning for simplicity.}
\label{tab:search_lr}
\end{table}

\section{Additional Results for Sequential Scaling of the Fine-tuned Models in Section \ref{section3}} \label{Appendix_scaling_ft}

Figure~\ref{fig:fig-qwen-scaling-merge}, \ref{fig:fig-s1-scaling-merge} provide pruned Qwen3-8B after Full FT and Lora FT on 1k data, pruned s1.1-7B after Full FT and LoRA FT on 1k data. Under SFT on 1k samples, we observe the same overall trend as in the 10k setting, but the improvements are consistently much smaller. In particular, while pruning 2 layers can still yield modest accuracy recovery at low thinking-token budgets, the gains quickly plateau and remain far below the unpruned models. As thinking tokens increase, the gap to the original model does not close and often becomes more pronounced, indicating that 1k-scale SFT likewise fails to restore sequential test-time scaling.

\begin{figure}[htbp]
\centering

\begin{subfigure}{\textwidth}
\centering
\begin{tikzpicture}
\begin{groupplot}[
  group style={group size=3 by 3, horizontal sep=1.2cm, vertical sep=0.6cm},
  width=0.31\textwidth, height=0.22\textwidth,
  xmode=log, log basis x=2,
  xtick={512,1024,2048,4096,8192},
  xticklabels={512,1024,2048,4096,8192},
  grid=major, grid style={dashed,gray!30},
  tick style={line width=0.6pt},
  label style={font=\small},
  tick label style={font=\small},
  xticklabel style={font=\scriptsize},
  title style={font=\bfseries\small},
  ylabel={Accuracy},
]
\nextgroupplot[title={MATH500}, ylabel={Acc (ShortGPT Full)}, legend to name=legendFull, legend columns=3, legend style={draw=none, font=\small}]
\addplot+[seriesA] coordinates {(512,0.875) (1024,0.905) (2048,0.924) (4096,0.929) (8192,0.960)}; \addlegendentry{No pruning}
\addplot+[seriesB] coordinates {(512,0.523) (1024,0.683) (2048,0.755) (4096,0.774) (8192,0.797)}; \addlegendentry{Prune 1 layer}
\addplot+[seriesC] coordinates {(512,0.525) (1024,0.639) (2048,0.685) (4096,0.718) (8192,0.732)}; \addlegendentry{Prune 2 layers}
\nextgroupplot[title={GPQA Diamond}, ylabel={}]
\addplot+[seriesA] coordinates {(512,0.446) (1024,0.441) (2048,0.478) (4096,0.530) (8192,0.564)};
\addplot+[seriesB] coordinates {(512,0.362) (1024,0.345) (2048,0.377) (4096,0.387) (8192,0.382)};
\addplot+[seriesC] coordinates {(512,0.320) (1024,0.337) (2048,0.351) (4096,0.402) (8192,0.406)};
\nextgroupplot[title={AIME24}, ylabel={}, ymin=0, ymax=0.7]
\addplot+[seriesA] coordinates {(512,0.556) (1024,0.422) (2048,0.489) (4096,0.467) (8192,0.667)};
\addplot+[seriesB] coordinates {(512,0.033) (1024,0.033) (2048,0.089) (4096,0.122) (8192,0.144)};
\addplot+[seriesC] coordinates {(512,0.044) (1024,0.033) (2048,0.044) (4096,0.100) (8192,0.122)};
\nextgroupplot[ylabel={Acc (Reverse Full)}]
\addplot+[seriesA] coordinates {(512,0.875) (1024,0.905) (2048,0.924) (4096,0.929) (8192,0.960)};
\addplot+[seriesB] coordinates {(512,0.676) (1024,0.749) (2048,0.807) (4096,0.839) (8192,0.847)};
\addplot+[seriesC] coordinates {(512,0.677) (1024,0.750) (2048,0.801) (4096,0.835) (8192,0.852)};
\nextgroupplot[ylabel={}]
\addplot+[seriesA] coordinates {(512,0.446) (1024,0.441) (2048,0.478) (4096,0.530) (8192,0.564)};
\addplot+[seriesB] coordinates {(512,0.381) (1024,0.394) (2048,0.418) (4096,0.442) (8192,0.426)};
\addplot+[seriesC] coordinates {(512,0.387) (1024,0.409) (2048,0.424) (4096,0.435) (8192,0.416)};
\nextgroupplot[ylabel={}, ymin=0, ymax=0.7]
\addplot+[seriesA] coordinates {(512,0.556) (1024,0.422) (2048,0.489) (4096,0.467) (8192,0.667)};
\addplot+[seriesB] coordinates {(512,0.111) (1024,0.156) (2048,0.133) (4096,0.222) (8192,0.211)};
\addplot+[seriesC] coordinates {(512,0.144) (1024,0.156) (2048,0.211) (4096,0.244) (8192,0.189)};
\nextgroupplot[ylabel={Acc (LaCo Full)}, xlabel={Thinking tokens}]
\addplot+[seriesA] coordinates {(512,0.875) (1024,0.905) (2048,0.924) (4096,0.929) (8192,0.960)};
\addplot+[seriesB] coordinates {(512,0.561) (1024,0.716) (2048,0.793) (4096,0.833) (8192,0.851)};
\addplot+[seriesC] coordinates {(512,0.545) (1024,0.681) (2048,0.742) (4096,0.789) (8192,0.791)};
\nextgroupplot[ylabel={}, xlabel={Thinking tokens}]
\addplot+[seriesA] coordinates {(512,0.446) (1024,0.441) (2048,0.478) (4096,0.530) (8192,0.564)};
\addplot+[seriesB] coordinates {(512,0.366) (1024,0.364) (2048,0.409) (4096,0.394) (8192,0.419)};
\addplot+[seriesC] coordinates {(512,0.337) (1024,0.372) (2048,0.371) (4096,0.386) (8192,0.403)};
\nextgroupplot[ylabel={}, xlabel={Thinking tokens}, ymin=0, ymax=0.7]
\addplot+[seriesA] coordinates {(512,0.556) (1024,0.422) (2048,0.489) (4096,0.467) (8192,0.667)};
\addplot+[seriesB] coordinates {(512,0.078) (1024,0.089) (2048,0.133) (4096,0.200) (8192,0.211)};
\addplot+[seriesC] coordinates {(512,0.056) (1024,0.078) (2048,0.100) (4096,0.133) (8192,0.167)};
\end{groupplot}
\end{tikzpicture}
\vspace{0.2em}
\pgfplotslegendfromname{legendFull}
\subcaption{Full Fine-tuning}
\end{subfigure}

\vspace{2em} 

\begin{subfigure}{\textwidth}
\centering
\begin{tikzpicture}
\begin{groupplot}[
  group style={group size=3 by 3, horizontal sep=1.2cm, vertical sep=0.6cm},
  width=0.31\textwidth, height=0.22\textwidth,
  xmode=log, log basis x=2,
  xtick={512,1024,2048,4096,8192},
  xticklabels={512,1024,2048,4096,8192},
  grid=major, grid style={dashed,gray!30},
  tick style={line width=0.6pt},
  label style={font=\small},
  tick label style={font=\small},
  xticklabel style={font=\scriptsize},
  title style={font=\bfseries\small},
  ylabel={Accuracy},
]
\nextgroupplot[title={MATH500}, ylabel={Acc (ShortGPT LoRA)}, legend to name=legendLoRA, legend columns=3, legend style={draw=none, font=\small}]
\addplot+[seriesA] coordinates {(512,0.875) (1024,0.905) (2048,0.924) (4096,0.929) (8192,0.960)}; \addlegendentry{No pruning}
\addplot+[seriesB] coordinates {(512,0.577) (1024,0.651) (2048,0.711) (4096,0.745) (8192,0.737)}; \addlegendentry{Prune 1 layer}
\addplot+[seriesC] coordinates {(512,0.509) (1024,0.603) (2048,0.639) (4096,0.696) (8192,0.701)}; \addlegendentry{Prune 2 layers}
\nextgroupplot[title={GPQA Diamond}, ylabel={}]
\addplot+[seriesA] coordinates {(512,0.446) (1024,0.441) (2048,0.478) (4096,0.530) (8192,0.564)};
\addplot+[seriesB] coordinates {(512,0.365) (1024,0.391) (2048,0.401) (4096,0.427) (8192,0.449)};
\addplot+[seriesC] coordinates {(512,0.306) (1024,0.358) (2048,0.320) (4096,0.362) (8192,0.386)};
\nextgroupplot[title={AIME24}, ylabel={}, ymin=0, ymax=0.72]
\addplot+[seriesA] coordinates {(512,0.556) (1024,0.422) (2048,0.489) (4096,0.467) (8192,0.667)};
\addplot+[seriesB] coordinates {(512,0.200) (1024,0.167) (2048,0.167) (4096,0.256) (8192,0.367)};
\addplot+[seriesC] coordinates {(512,0.033) (1024,0.022) (2048,0.022) (4096,0.067) (8192,0.144)};
\nextgroupplot[
  ylabel={Acc (Reverse LoRA)},
  legend to name=legendQwenMergeReverse,
  legend columns=3,
  legend style={draw=none, /tikz/every even column/.style={column sep=0.8em}, font=\small}
]
\addplot+[seriesA] coordinates {(512,0.875) (1024,0.905) (2048,0.924) (4096,0.929) (8192,0.960)}; \addlegendentry{No pruning}
\addplot+[seriesB] coordinates {(512,0.606) (1024,0.687) (2048,0.748) (4096,0.778) (8192,0.799)}; \addlegendentry{Prune 1 layer}
\addplot+[seriesC] coordinates {(512,0.634) (1024,0.677) (2048,0.697) (4096,0.711) (8192,0.724)}; \addlegendentry{Prune 2 layer}
\nextgroupplot[ylabel={}]
\addplot+[seriesA] coordinates {(512,0.446) (1024,0.441) (2048,0.478) (4096,0.530) (8192,0.564)};
\addplot+[seriesB] coordinates {(512,0.375) (1024,0.427) (2048,0.424) (4096,0.460) (8192,0.468)};
\addplot+[seriesC] coordinates {(512,0.345) (1024,0.377) (2048,0.421) (4096,0.379) (8192,0.407)};
\nextgroupplot[ylabel={}, ymin=0, ymax=0.72]
\addplot+[seriesA] coordinates {(512,0.556) (1024,0.422) (2048,0.489) (4096,0.467) (8192,0.667)};
\addplot+[seriesB] coordinates {(512,0.044) (1024,0.089) (2048,0.122) (4096,0.200) (8192,0.156)};
\addplot+[seriesC] coordinates {(512,0.089) (1024,0.111) (2048,0.122) (4096,0.156) (8192,0.167)};
\nextgroupplot[
  ylabel={Acc (LaCo LoRA)},
  xlabel={Thinking tokens},
  legend to name=legendQwenMergeLaCo,
  legend columns=3,
  legend style={draw=none, /tikz/every even column/.style={column sep=0.8em}, font=\small}
]
\addplot+[seriesA] coordinates {(512,0.875) (1024,0.905) (2048,0.924) (4096,0.929) (8192,0.960)}; \addlegendentry{No pruning}
\addplot+[seriesB] coordinates {(512,0.557) (1024,0.642) (2048,0.703) (4096,0.761) (8192,0.812)}; \addlegendentry{Prune 1 layer}
\addplot+[seriesC] coordinates {(512,0.524) (1024,0.614) (2048,0.663) (4096,0.721) (8192,0.752)}; \addlegendentry{Prune 2 layer}
\nextgroupplot[ylabel={}, xlabel={Thinking tokens}]
\addplot+[seriesA] coordinates {(512,0.446) (1024,0.441) (2048,0.478) (4096,0.530) (8192,0.564)};
\addplot+[seriesB] coordinates {(512,0.396) (1024,0.382) (2048,0.441) (4096,0.471) (8192,0.498)};
\addplot+[seriesC] coordinates {(512,0.357) (1024,0.354) (2048,0.384) (4096,0.406) (8192,0.414)};
\nextgroupplot[ylabel={}, ymin=0, ymax=0.72]
\addplot+[seriesA] coordinates {(512,0.556) (1024,0.422) (2048,0.489) (4096,0.467) (8192,0.667)};
\addplot+[seriesB] coordinates {(512,0.089) (1024,0.122) (2048,0.144) (4096,0.178) (8192,0.144)};
\addplot+[seriesC] coordinates {(512,0.044) (1024,0.067) (2048,0.078) (4096,0.100) (8192,0.167)};
\end{groupplot}
\end{tikzpicture}
\vspace{0.2em}
\pgfplotslegendfromname{legendLoRA}
\subcaption{LoRA Fine-tuning}
\end{subfigure}

\caption{Comparison of test-time scaling for Qwen3-8B under Full FT and LoRA FT across different pruning methods and depths.}
\label{fig:fig-qwen-scaling-merge}
\end{figure}
\begin{figure*}[htbp]
\centering

\begin{subfigure}{\textwidth}
\centering
\begin{tikzpicture}
\begin{groupplot}[
  group style={group size=3 by 3, horizontal sep=1.2cm, vertical sep=0.6cm},
  width=0.31\textwidth, height=0.22\textwidth,
  xmode=log, log basis x=2,
  xtick={512,1024,2048,4096,8192},
  xticklabels={512,1024,2048,4096,8192},
  grid=major, grid style={dashed,gray!30},
  tick style={line width=0.6pt},
  label style={font=\small},
  tick label style={font=\small},
  xticklabel style={font=\scriptsize},
  title style={font=\bfseries\small},
  ylabel={Accuracy},
  scaled y ticks=false, 
  y tick label style={
    /pgf/number format/fixed,
    /pgf/number format/precision=2, 
    /pgf/number format/fixed zerofill 
  }
]

\nextgroupplot[
    title={MATH500}, 
    ylabel={Acc (ShortGPT Full)}, 
    legend to name=legendS1Full, 
    legend columns=3, 
    legend style={draw=none, font=\small, /tikz/every even column/.style={column sep=0.8em}}
]
\addplot+[seriesA] coordinates {(512,0.626) (1024,0.701) (2048,0.760) (4096,0.783) (8192,0.803)}; \addlegendentry{No pruning}
\addplot+[seriesB] coordinates {(512,0.462) (1024,0.532) (2048,0.554) (4096,0.580) (8192,0.590)}; \addlegendentry{Prune 1 layer}
\addplot+[seriesC] coordinates {(512,0.418) (1024,0.455) (2048,0.473) (4096,0.495) (8192,0.477)}; \addlegendentry{Prune 2 layers}

\nextgroupplot[title={GPQA Diamond}, ylabel={}]
\addplot+[seriesA] coordinates {(512,0.390) (1024,0.353) (2048,0.401) (4096,0.411) (8192,0.428)};
\addplot+[seriesB] coordinates {(512,0.341) (1024,0.352) (2048,0.334) (4096,0.332) (8192,0.308)};
\addplot+[seriesC] coordinates {(512,0.320) (1024,0.333) (2048,0.313) (4096,0.318) (8192,0.295)};

\nextgroupplot[title={AIME24}, ylabel={}]
\addplot+[seriesA] coordinates {(512,0.044) (1024,0.067) (2048,0.133) (4096,0.089) (8192,0.144)};
\addplot+[seriesB] coordinates {(512,0.011) (1024,0.022) (2048,0.022) (4096,0.033) (8192,0.022)};
\addplot+[seriesC] coordinates {(512,0.000) (1024,0.011) (2048,0.000) (4096,0.011) (8192,0.000)};

\nextgroupplot[ylabel={Acc (Reverse Full)}]
\addplot+[seriesA] coordinates {(512,0.626) (1024,0.701) (2048,0.760) (4096,0.783) (8192,0.803)};
\addplot+[seriesB] coordinates {(512,0.492) (1024,0.588) (2048,0.641) (4096,0.650) (8192,0.665)};
\addplot+[seriesC] coordinates {(512,0.467) (1024,0.563) (2048,0.593) (4096,0.610) (8192,0.609)};

\nextgroupplot[ylabel={}]
\addplot+[seriesA] coordinates {(512,0.390) (1024,0.353) (2048,0.401) (4096,0.411) (8192,0.428)};
\addplot+[seriesB] coordinates {(512,0.350) (1024,0.305) (2048,0.342) (4096,0.325) (8192,0.330)};
\addplot+[seriesC] coordinates {(512,0.341) (1024,0.335) (2048,0.326) (4096,0.311) (8192,0.293)};

\nextgroupplot[ylabel={}]
\addplot+[seriesA] coordinates {(512,0.044) (1024,0.067) (2048,0.133) (4096,0.089) (8192,0.144)};
\addplot+[seriesB] coordinates {(512,0.011) (1024,0.033) (2048,0.044) (4096,0.044) (8192,0.044)};
\addplot+[seriesC] coordinates {(512,0.011) (1024,0.000) (2048,0.033) (4096,0.044) (8192,0.044)};

\nextgroupplot[ylabel={Acc (LaCo Full)}, xlabel={Thinking tokens}]
\addplot+[seriesA] coordinates {(512,0.626) (1024,0.701) (2048,0.760) (4096,0.783) (8192,0.803)};
\addplot+[seriesB] coordinates {(512,0.395) (1024,0.480) (2048,0.518) (4096,0.539) (8192,0.541)};
\addplot+[seriesC] coordinates {(512,0.356) (1024,0.422) (2048,0.459) (4096,0.465) (8192,0.458)};

\nextgroupplot[ylabel={}, xlabel={Thinking tokens}]
\addplot+[seriesA] coordinates {(512,0.390) (1024,0.353) (2048,0.401) (4096,0.411) (8192,0.428)};
\addplot+[seriesB] coordinates {(512,0.331) (1024,0.321) (2048,0.323) (4096,0.337) (8192,0.308)};
\addplot+[seriesC] coordinates {(512,0.298) (1024,0.298) (2048,0.295) (4096,0.279) (8192,0.280)};

\nextgroupplot[ylabel={}, xlabel={Thinking tokens}]
\addplot+[seriesA] coordinates {(512,0.044) (1024,0.067) (2048,0.133) (4096,0.089) (8192,0.144)};
\addplot+[seriesB] coordinates {(512,0.000) (1024,0.000) (2048,0.033) (4096,0.011) (8192,0.011)};
\addplot+[seriesC] coordinates {(512,0.000) (1024,0.000) (2048,0.011) (4096,0.000) (8192,0.011)};
\end{groupplot}
\end{tikzpicture}

\vspace{0.4em}
\pgfplotslegendfromname{legendS1Full} 

\subcaption{s1.1-7B Full Fine-tuning}
\end{subfigure}

\begin{subfigure}{\textwidth}
\centering
\begin{tikzpicture}
\begin{groupplot}[
  group style={group size=3 by 3, horizontal sep=1.2cm, vertical sep=0.6cm},
  width=0.31\textwidth, height=0.22\textwidth,
  xmode=log, log basis x=2,
  xtick={512,1024,2048,4096,8192},
  xticklabels={512,1024,2048,4096,8192},
  grid=major, grid style={dashed,gray!30},
  tick style={line width=0.6pt},
  label style={font=\small},
  tick label style={font=\small},
  xticklabel style={font=\scriptsize},
  title style={font=\bfseries\small},
  ylabel={Accuracy},
  scaled y ticks=false,
  y tick label style={/pgf/number format/fixed, /pgf/number format/precision=2}
]
\nextgroupplot[
    title={MATH500}, 
    ylabel={Acc (ShortGPT LoRA)}, 
    legend to name=legendS1LoRA, 
    legend columns=3, 
    legend style={draw=none, fill=none, font=\small}
]
\addplot+[seriesA] coordinates {(512,0.626) (1024,0.701) (2048,0.760) (4096,0.783) (8192,0.803)}; \addlegendentry{No pruning}
\addplot+[seriesB] coordinates {(512,0.481) (1024,0.565) (2048,0.589) (4096,0.629) (8192,0.632)}; \addlegendentry{Prune 1 layer}
\addplot+[seriesC] coordinates {(512,0.411) (1024,0.459) (2048,0.489) (4096,0.514) (8192,0.507)}; \addlegendentry{Prune 2 layers}

\nextgroupplot[title={GPQA Diamond}, ylabel={}]
\addplot+[seriesA] coordinates {(512,0.390) (1024,0.353) (2048,0.401) (4096,0.411) (8192,0.428)};
\addplot+[seriesB] coordinates {(512,0.354) (1024,0.357) (2048,0.337) (4096,0.343) (8192,0.335)};
\addplot+[seriesC] coordinates {(512,0.311) (1024,0.288) (2048,0.258) (4096,0.305) (8192,0.313)};

\nextgroupplot[title={AIME24}, ylabel={}, ymin=0, ymax=0.16, ytick={0,0.04,0.08,0.12,0.16}]
\addplot+[seriesA] coordinates {(512,0.044) (1024,0.067) (2048,0.133) (4096,0.089) (8192,0.144)};
\addplot+[seriesB] coordinates {(512,0.000) (1024,0.011) (2048,0.033) (4096,0.067) (8192,0.067)};
\addplot+[seriesC] coordinates {(512,0.000) (1024,0.011) (2048,0.022) (4096,0.000) (8192,0.011)};

\nextgroupplot[ylabel={Acc (Reverse LoRA)}]
\addplot+[seriesA] coordinates {(512,0.626) (1024,0.701) (2048,0.760) (4096,0.783) (8192,0.803)};
\addplot+[seriesB] coordinates {(512,0.414) (1024,0.467) (2048,0.483) (4096,0.481) (8192,0.467)};
\addplot+[seriesC] coordinates {(512,0.002) (1024,0.005) (2048,0.006) (4096,0.047) (8192,0.057)};

\nextgroupplot[ylabel={}]
\addplot+[seriesA] coordinates {(512,0.390) (1024,0.353) (2048,0.401) (4096,0.411) (8192,0.428)};
\addplot+[seriesB] coordinates {(512,0.299) (1024,0.317) (2048,0.310) (4096,0.341) (8192,0.321)};
\addplot+[seriesC] coordinates {(512,0.254) (1024,0.242) (2048,0.268) (4096,0.247) (8192,0.262)};

\nextgroupplot[ylabel={}, ymin=0, ymax=0.16, ytick={0,0.04,0.08,0.12,0.16}]
\addplot+[seriesA] coordinates {(512,0.044) (1024,0.067) (2048,0.133) (4096,0.089) (8192,0.144)};
\addplot+[seriesB] coordinates {(512,0.011) (1024,0.011) (2048,0.022) (4096,0.011) (8192,0.033)};
\addplot+[seriesC] coordinates {(512,0.000) (1024,0.000) (2048,0.000) (4096,0.011) (8192,0.011)};

\nextgroupplot[ylabel={Acc (LaCo LoRA)}, xlabel={Thinking tokens}]
\addplot+[seriesA] coordinates {(512,0.626) (1024,0.701) (2048,0.760) (4096,0.783) (8192,0.803)};
\addplot+[seriesB] coordinates {(512,0.423) (1024,0.485) (2048,0.521) (4096,0.531) (8192,0.542)};
\addplot+[seriesC] coordinates {(512,0.333) (1024,0.397) (2048,0.429) (4096,0.455) (8192,0.450)};

\nextgroupplot[ylabel={}, xlabel={Thinking tokens}]
\addplot+[seriesA] coordinates {(512,0.390) (1024,0.353) (2048,0.401) (4096,0.411) (8192,0.428)};
\addplot+[seriesB] coordinates {(512,0.296) (1024,0.290) (2048,0.323) (4096,0.301) (8192,0.310)};
\addplot+[seriesC] coordinates {(512,0.313) (1024,0.331) (2048,0.281) (4096,0.291) (8192,0.271)};

\nextgroupplot[ylabel={}, xlabel={Thinking tokens}, ymin=0, ymax=0.16, ytick={0,0.04,0.08,0.12,0.16}]
\addplot+[seriesA] coordinates {(512,0.044) (1024,0.067) (2048,0.133) (4096,0.089) (8192,0.144)};
\addplot+[seriesB] coordinates {(512,0.000) (1024,0.011) (2048,0.011) (4096,0.022) (8192,0.011)};
\addplot+[seriesC] coordinates {(512,0.000) (1024,0.000) (2048,0.000) (4096,0.022) (8192,0.000)};
\end{groupplot}
\end{tikzpicture}
\vspace{0.2em}
\pgfplotslegendfromname{legendS1LoRA}
\subcaption{s1.1-7B LoRA Fine-tuning}
\end{subfigure}

\caption{Sequential test-time scaling of s1.1-7B under Full FT and LoRA FT with different pruning depths.}
\label{fig:fig-s1-scaling-merge}
\end{figure*}

\newpage

\section{Results for Diverse Temperature, Repetition Penalty and Frequency Penalty}
\label{decoding}

Table~\ref{tab:aime24_temp_pruned_qwen3_8b}, \ref{tab:aime24_rep_pruned_qwen3_8b}, \ref{tab:aime24_freq_pruned_qwen3_8b} show the impact of temperature, repetition penalty and frequency penalty. Overall, varying temperature and applying penalty-based sampling fails to recover the test-time scaling losses caused by pruning. At best, moderate temperatures yield only marginal improvements, whereas stronger stochasticity or repetition penalties frequently degrade performance.

\begin{table}[h]
\centering

\begin{subtable}[t]{0.48\linewidth}
\centering
\small 
\resizebox{\linewidth}{!}{%
\begin{tabular}{lccccc}
\toprule
\textbf{Temp.} & \textbf{512} & \textbf{1024} & \textbf{2048} & \textbf{4096} & \textbf{8192} \\
\midrule
0.0 & 0.1889 & 0.2001 & 0.1965 & 0.2557 & 0.3630 \\
0.4 & 0.2556 & 0.2667 & 0.2812 & 0.3333 & 0.4297 \\
0.7 & 0.3148 & 0.2667 & 0.3333 & 0.3778 & 0.4815 \\
1.0 & 0.3519 & 0.3185 & 0.3593 & 0.3741 & 0.4481 \\
1.5 & 0.1481 & 0.1519 & 0.1630 & 0.2260 & 0.2260 \\
2.0 & 0.0074 & 0.0148 & 0.0148 & 0.0074 & 0.0037 \\
3.0 & 0.0000 & 0.0000 & 0.0000 & 0.0037 & 0.0000 \\
\bottomrule
\end{tabular}
}
\caption{1-layer pruned.}
\label{tab:aime24_temp_1layer}
\end{subtable}
\hfill
\begin{subtable}[t]{0.48\linewidth}
\centering
\small
\resizebox{\linewidth}{!}{%
\begin{tabular}{lccccc}
\toprule
\textbf{Temp.} & \textbf{512} & \textbf{1024} & \textbf{2048} & \textbf{4096} & \textbf{8192} \\
\midrule
0.0 & 0.0222 & 0.0296 & 0.0519 & 0.0852 & 0.0852 \\
0.4 & 0.0556 & 0.0815 & 0.0593 & 0.1074 & 0.1333 \\
0.7 & 0.0852 & 0.0890 & 0.0852 & 0.1000 & 0.1222 \\
1.0 & 0.0778 & 0.1074 & 0.1074 & 0.1222 & 0.1481 \\
1.5 & 0.0185 & 0.0111 & 0.0259 & 0.0222 & 0.0148 \\
2.0 & 0.0037 & 0.0000 & 0.0037 & 0.0000 & 0.0000 \\
3.0 & 0.0000 & 0.0000 & 0.0000 & 0.0000 & 0.0000 \\
\bottomrule
\end{tabular}
}
\caption{2-layer pruned.}
\label{tab:aime24_temp_2layer}
\end{subtable}

\caption{Results on AIME24 of pruned Qwen3-8B for sequential test-time scaling w.r.t. different temperature.}
\label{tab:aime24_temp_pruned_qwen3_8b}
\end{table}

\begin{table}[h]
\centering

\begin{subtable}[t]{0.495\linewidth}
\centering
\resizebox{\linewidth}{!}{
\begin{tabular}{lccccc}
\toprule
\textbf{Repetition} & \textbf{512} & \textbf{1024} & \textbf{2048} & \textbf{4096} & \textbf{8192} \\
\midrule
1.0 & 0.3444 & 0.3444 & 0.3444 & 0.4444 & 0.4778 \\
1.1 & 0.2556 & 0.2222 & 0.2444 & 0.3667 & 0.4667 \\
1.2 & 0.1889 & 0.1556 & 0.1889 & 0.2444 & 0.3667 \\
    1.5 & 0.0667 & 0.1222 & 0.1556 & 0.1111 & 0.1222 \\
\bottomrule
\end{tabular}}
\caption{1-layer pruned.}
\label{tab:aime24_rep_1layer}
\end{subtable}
\hfill
\begin{subtable}[t]{0.495\linewidth}
\centering
\resizebox{\linewidth}{!}{
\begin{tabular}{lccccc}
\toprule
\textbf{Repetition} & \textbf{512} & \textbf{1024} & \textbf{2048} & \textbf{4096} & \textbf{8192} \\
\midrule
1.0 & 0.0556 & 0.1667 & 0.1444 & 0.1000 & 0.1444 \\
1.1 & 0.0778 & 0.1000 & 0.0556 & 0.1000 & 0.1111 \\
1.2 & 0.0556 & 0.0333 & 0.0444 & 0.0333 & 0.0111 \\
    1.5 & 0.0000 & 0.0000 & 0.0000 & 0.0000 & 0.0000 \\
\bottomrule
\end{tabular}}
\caption{2-layer pruned.}
\label{tab:aime24_rep_2layer}
\end{subtable}

\caption{Results on AIME24 of pruned Qwen3-8B for sequential test-time scaling w.r.t. repetition penalty.}
\label{tab:aime24_rep_pruned_qwen3_8b}
\end{table}

\begin{table}[h]
\centering

\setlength{\tabcolsep}{4pt}

\begin{subtable}[t]{0.485\linewidth}
\centering
\resizebox{\linewidth}{!}{
\begin{tabular}{lccccc}
\toprule
\textbf{Frequency} & \textbf{512} & \textbf{1024} & \textbf{2048} & \textbf{4096} & \textbf{8192} \\
\midrule
0.0 & 0.3333 & 0.3333 & 0.3667 & 0.4333 & 0.4333 \\
0.1 & 0.0444 & 0.0333 & 0.0889 & 0.1444 & 0.1556 \\
0.3 & 0.0111 & 0.0333 & 0.0444 & 0.0667 & 0.0667 \\
0.5 & 0.0111 & 0.0222 & 0.0222 & 0.0333 & 0.0333 \\
0.7 & 0.0111 & 0.0111 & 0.0111 & 0.0222 & 0.0444 \\
    0.9 & 0.0111 & 0.0111 & 0.0222 & 0.0111 & 0.0111 \\
\bottomrule
\end{tabular}}
\caption{1-layer pruned.}
\label{tab:aime24_freq_1layer}
\end{subtable}
\hfill
\begin{subtable}[t]{0.485\linewidth}
\centering
\resizebox{\linewidth}{!}{
\begin{tabular}{lccccc}
\toprule
\textbf{Frequency} & \textbf{512} & \textbf{1024} & \textbf{2048} & \textbf{4096} & \textbf{8192} \\
\midrule
0.0 & 0.0778 & 0.2000 & 0.1667 & 0.1778 & 0.1778 \\
0.1 & 0.0111 & 0.0111 & 0.0333 & 0.0222 & 0.0444 \\
0.3 & 0.0333 & 0.0111 & 0.0111 & 0.0111 & 0.0222 \\
0.5 & 0.0000 & 0.0111 & 0.0222 & 0.0222 & 0.0111 \\
0.7 & 0.0000 & 0.0111 & 0.0111 & 0.0000 & 0.0222 \\
    0.9 & 0.0111 & 0.0000 & 0.0111 & 0.0000 & 0.0000 \\
\bottomrule
\end{tabular}}
\caption{2-layer pruned.}
\label{tab:aime24_freq_2layer}
\end{subtable}

\caption{Results on AIME24 of pruned Qwen3-8B for sequential test-time scaling w.r.t. frequency penalty.}
\label{tab:aime24_freq_pruned_qwen3_8b}
\end{table}



\section{Brute-force Layer Ablation for Sequential Test-time Scaling} \label{Appendix_brute_force}

\begin{figure}[htbp]
\centering
\includegraphics[width=1\textwidth, trim=80 130 80 135, clip]{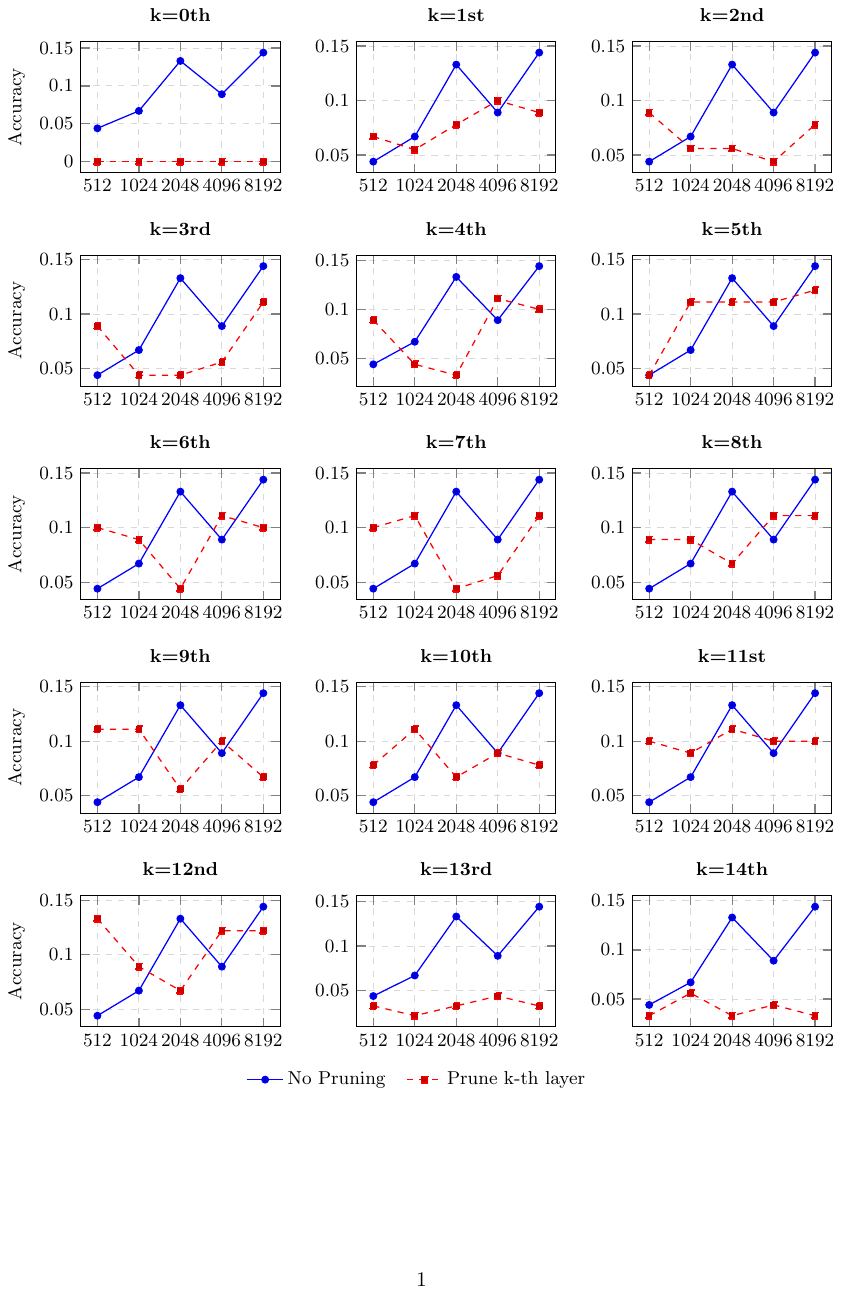}
\caption{Brute-force of s1.1-7B layer ablation for sequential test-time scaling (Part 1)}
\label{fig:brute-force1}
\end{figure}

\begin{figure}[htbp]
\centering
\includegraphics[width=1\textwidth, trim=80 130 80 135, clip]{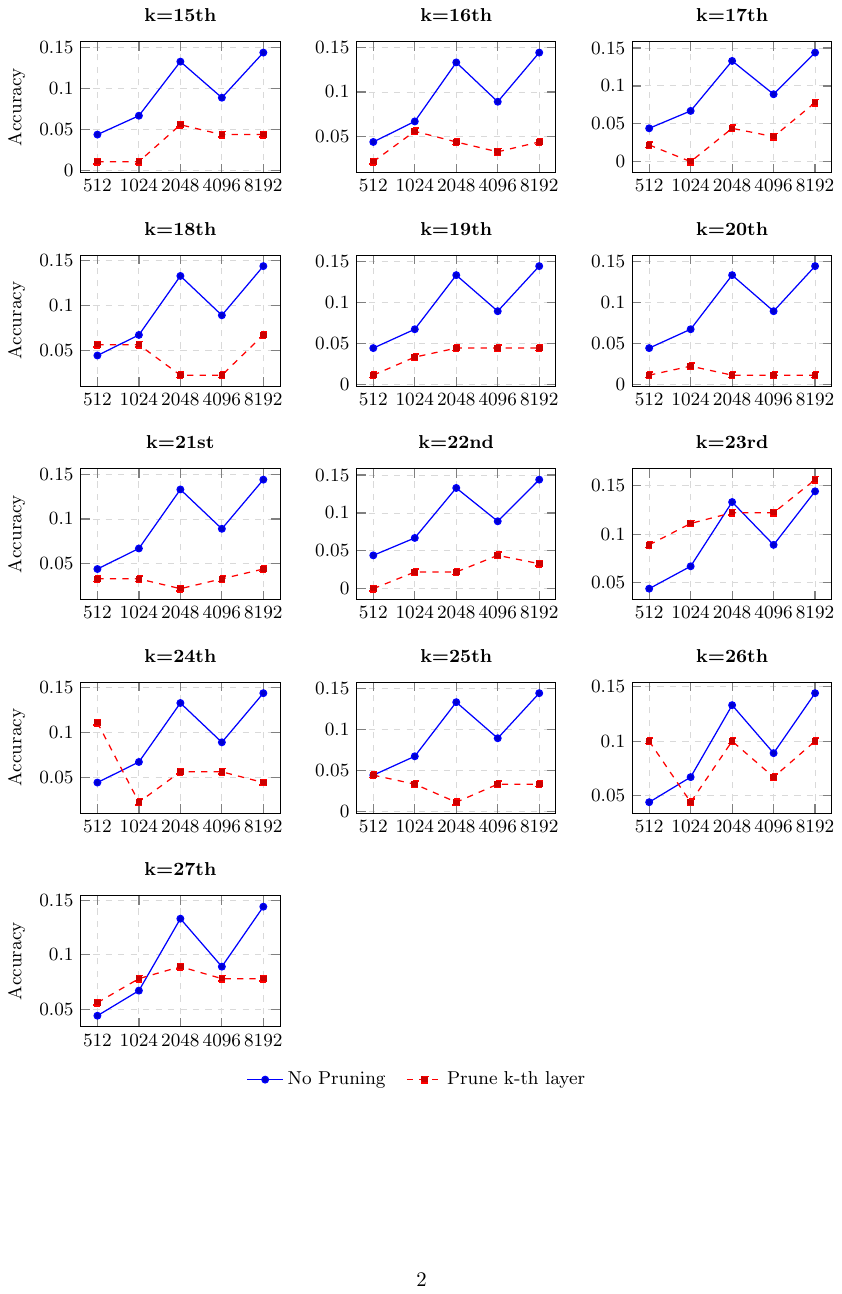}
\caption{Brute-force of s1.1-7B layer ablation for sequential test-time scaling (Part 2)}
\label{fig:brute-force2}
\end{figure}

\begin{figure}[htbp]
\centering
\includegraphics[width=1\textwidth, trim=80 100 80 100, clip]{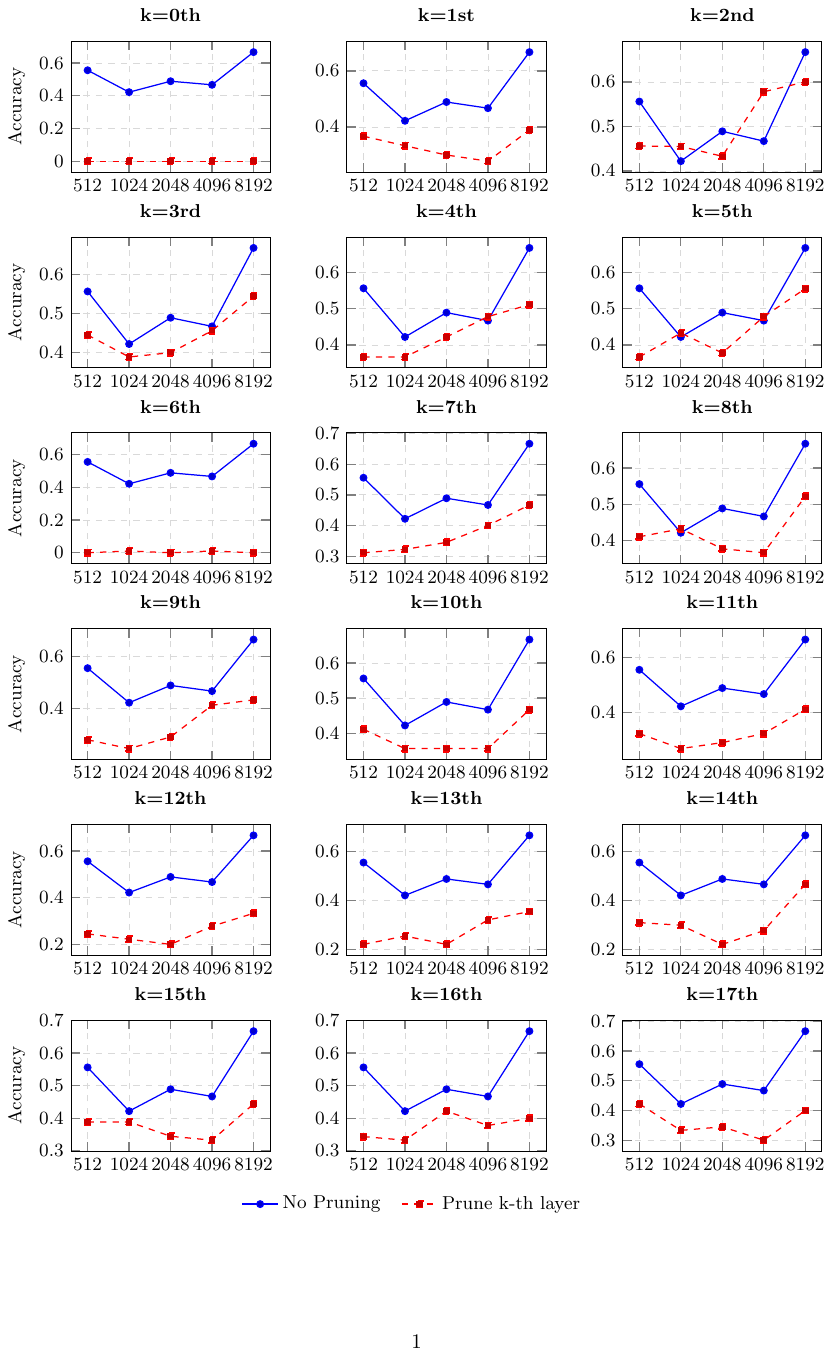}
\caption{Brute-force of Qwen3-8B layer ablation for sequential test-time scaling (Part 1)}
\label{fig:brute-force3}
\end{figure}

\begin{figure}[htbp]
\centering
\includegraphics[width=1\textwidth, trim=80 100 80 100, clip]{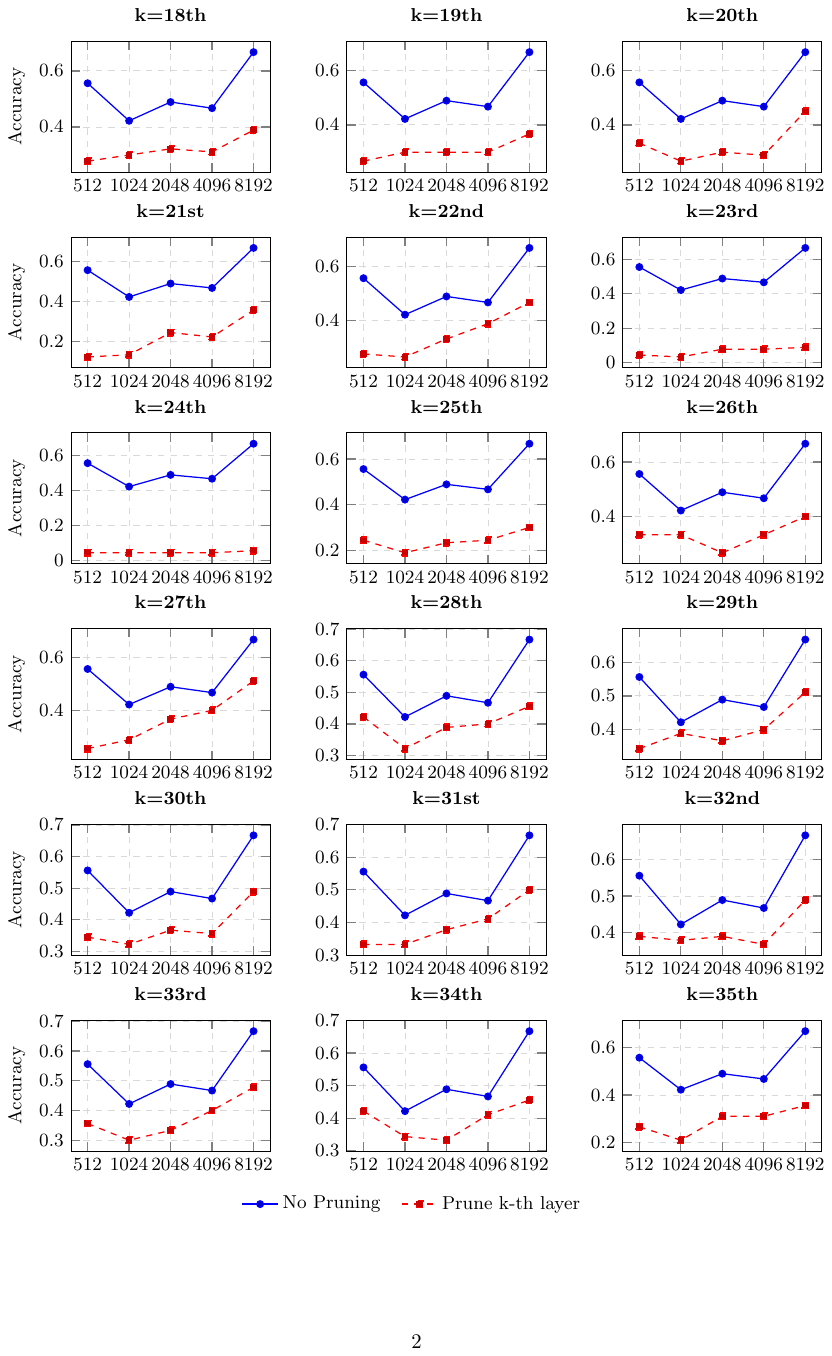}
\caption{Brute-force of Qwen3-8B layer ablation for sequential test-time scaling (Part 2)}
\label{fig:brute-force4}
\end{figure}

We prune one layer at a time and evaluate the sequential test-time scaling of the corresponding variants, as shown in Figure~\ref{fig:brute-force1}, \ref{fig:brute-force2}, \ref{fig:brute-force3} and \ref{fig:brute-force4}. Most layers make significant contributions to test-time scaling, and removing even a single layer can substantially harm this ability.

\section{SFT Datasets}
\label{sec:datasets}

Our experiments primarily utilize the \textbf{s1K-1.1} dataset\footnote{\url{https://huggingface.co/datasets/simplescaling/s1K-1.1}}, a carefully curated collection of 1,000 high-quality reasoning questions. These problems are selected from an initial pool of 59,000 examples spanning 51 scientific and mathematical domains, including geometry, number theory, combinatorics, physics, and theoretical computer science. Each question is paired with a \textbf{reasoning trace distilled from Gemini Flash Thinking API} \cite{comanici2025gemini}, accumulating to a total of 4.7 million tokens of step-by-step thought data. The final 1,000 samples were chosen through a rigorous three-stage filtering pipeline that prioritizes \textit{Quality} (removing malformed items), \textit{Difficulty} (favoring long reasoning chains), and \textit{Diversity} (using weighted sampling across domains). Despite its compact size, s1K provides a highly sample-efficient training source for teaching complex, sequential reasoning.

To further evaluate the impact of data scale and distribution, we additionally employ a subset of the \textbf{OpenR1-Math-220k} dataset\footnote{\url{https://huggingface.co/datasets/open-r1/OpenR1-Math-220k}}. This large-scale mathematical reasoning benchmark contains 220,000 problems from NuminaMath 1.5, each paired with 2--4 long-form reasoning traces generated by DeepSeek R1. The traces undergo rigorous validation, primarily via Math Verify, with a subset adjudged by Llama-3.3-70B-Instruct to ensure correctness. For our expanded evaluation, we randomly sampled 10,000 examples from this dataset and applied Full Supervised Fine-Tuning (SFT) to the pruned s1.1-7B model. maintaining the same hyperparameter settings as the s1 experiments (learning rate of $1\text{e-}5$). Results indicate that while increasing the dataset size by $10\times$ substantially improves SFT effectiveness, it still only partially restores the test-time scaling performance of models where two layers have been pruned.

\section{Additional Qualitative Examples of Repetitive Reasoning} \label{Appendix_qualitative_example}

We provide additional qualitative examples of the repetitive reasoning phenomenon in pruned models in Tables~\ref{tab:s1.1_aime24}, extending the discussion presented in Section~\ref{Section_5.1}.

\begin{longtable}{@{}p{\columnwidth}@{}}
\caption{{\textbf{Failure Analysis of a Pruned s1.1-7B Model on an AIME24 Problem.} The table shows the reference solution (\textcolor{olmoeDarkYellow}{yellow}) against the model's generated reasoning (\textcolor{defaultlightblue}{light blue}) and answer (\textcolor{defaultblue}{blue}). The reasoning trace highlights a failure where the model makes a calculation error and enters a self-contradictory loop.}} \\
\toprule
Let $p$ be the least prime number for which there exists a positive integer $n$ such that $n^{4}+1$ is divisible by $p^{2}$. Find the least positive integer $m$ such that $m^{4}+1$ is divisible by $p^{2}$.

\\
\defaultyellow{
If $p=2$, then $4\mid n^4+1$ for some integer $n$. But $\left(n^2\right)^2\equiv0$ or $1\pmod4$, so it is impossible. Thus $p$ is an odd prime.
For integer $n$ such that $p^2\mid n^4+1$, we have $p\mid n^4+1$, hence $p\nmid n^4-1$, but $p\mid n^8-1$. By [Fermat's Little Theorem](https://artofproblemsolving.com/wiki/index.php/Fermat
\begin{equation*}
p\mid\gcd\left(n^{p-1}-1,n^8-1\right)=n^{\gcd(p-1,8)}-1.
\end{equation*}
Here, $\gcd(p-1,8)$ mustn't be divide into $4$ or otherwise $p\mid n^{\gcd(p-1,8)}-1\mid n^4-1$, which contradicts. So $\gcd(p-1,8)=8$, and so $8\mid p-1$. The smallest such prime is clearly $p=17=2\times8+1$.
So we have to find the smallest positive integer $m$ such that $17\mid m^4+1$. We first find the remainder of $m$ divided by $17$ by doing
\[
\begin{array}{|c|cccccccccccccccc|}
\hline
\vphantom{\tfrac11}x \pmod{17} & 1 & 2 & 3 & 4 & 5 & 6 & 7 & 8 & 9 & 10 & 11 & 12 & 13 & 14 & 15 & 16 \\
\hline
\vphantom{\dfrac11}x^4+1 \pmod{17} & 2 & 0 & 14 & 2 & 14 & 5 & 5 & 0 & 0 & 5 & 5 & 14 & 2 & 14 & 0 & 2 \\
\hline
\end{array}
\]
So $m\equiv\pm2$, $\pm8\pmod{17}$. If $m\equiv2\pmod{17}$, let $m=17k+2$, by the binomial theorem,
\begin{align*}
0&\equiv(17k+2)^4+1\equiv\binom{4}{1}(17k)(2)^3+2^4+1=17(1+32k)\pmod{17^2}\\[3pt]
\implies0&\equiv1+32k\equiv1-2k\pmod{17}.
\end{align*}
So the smallest possible $k=9$, and $m=155$.
If $m\equiv-2\pmod{17}$, let $m=17k-2$, by the binomial theorem,
\begin{align*}
0&\equiv(17k-2)^4+1\equiv\binom{4}{1}(17k)(-2)^3+2^4+1=17(1-32k)\pmod{17^2}\\[3pt]
\implies0&\equiv1-32k\equiv1+2k\pmod{17}.
\end{align*}
So the smallest possible $k=8$, and $m=134$.
If $m\equiv8\pmod{17}$, let $m=17k+8$, by the binomial theorem,
\begin{align*}
0&\equiv(17k+8)^4+1\equiv\binom{4}{1}(17k)(8)^3+8^4+1=17(241+2048k)\pmod{17^2}\\[3pt]
\implies0&\equiv241+2048k\equiv3+8k\pmod{17}.
\end{align*}
}
\\
\defaultyellow{
So the smallest possible $k=6$, and $m=110$.
If $m\equiv-8\pmod{17}$, let $m=17k-8$, by the binomial theorem,
\begin{align*}
0&\equiv(17k-8)^4+1\equiv\binom{4}{1}(17k)(-8)^3+8^4+1=17(241-2048k)\pmod{17^2}\\[3pt]
\implies0&\equiv241+2048k\equiv3+9k\pmod{17}.
\end{align*}
So the smallest possible $k=11$, and $m=179$.
In conclusion, the smallest possible $m$ is $\boxed{110}$.
Solution by Quantum-Phantom
We work in the ring $\mathbb Z/289\mathbb Z$ and use the formula
\[\sqrt[4]{-1}=\pm\sqrt{\frac12}\pm\sqrt{-\frac12}.\]
Since $-\frac12=144$, the expression becomes $\pm12\pm12i$, and it is easily calculated via Hensel that $i=38$, thus giving an answer of $\boxed{110}$.
Note that $n^4 + 1 \equiv 0 \pmod{p}$ means $\text{ord}_{p}(n) = 8 \mid p-1.$ The smallest prime that does this is $17$ and $2^4 + 1 = 17$ for example. Now let $g$ be a primitive root of $17^2.$ The satisfying $n$ are of the form, $g^{\frac{p(p-1)}{8}}, g^{3\frac{p(p-1)}{8}}, g^{5\frac{p(p-1)}{8}}, g^{7\frac{p(p-1)}{8}}.$ So if we find one such $n$, then all $n$ are $n, n^3, n^5, n^7.$ Consider the $2$ from before. Note $17^2 \mid 2^{4 \cdot 17} + 1$ by LTE. Hence the possible $n$ are, $2^{17}, 2^{51}, 2^{85}, 2^{119}.$ Some modular arithmetic yields that $2^{51} \equiv \boxed{110}$ is the least value.
}
\\
\textbf{[Repetitive generation with 8192 thinking tokens]}
\\
\defaultlightblue{
Okay, so I need to find the least prime number $p$ such that there exists a positive integer $n$ where $p^2$ divides $n^4 + 1$. Then, once I find that $p$, I have to find the smallest positive integer $m$ such that $m^4 + 1$ is divisible by $p^2$.

First, let me recall some number theory concepts that might help here. I remember that for primes $p$, if $p$ divides $n^4 + 1$, then $n^4 \equiv -1 \pmod{p}$. That means $n^8 \equiv 1 \pmod{p}$, right? Because if $n^4 \equiv -1 \pmod{p}$, then squaring both sides gives $n^8 \equiv 1 \pmod{p}$. So the order of $n$ modulo $p$ divides 8. But primes $p$ for which there exists an integer $n$ with $n^4 \equiv -1 \pmod{p}$ must have that 8 is the order. Wait, not necessarily 8. Wait, maybe I need to think again.

Wait, if $n^4 \equiv -1 \pmod{p}$, then $n^8 \equiv 1 \pmod{p}$. So the order of $n$ modulo $p$ must divide 8. So possible orders are divisors of 8: 1, 2, 4, 8. But if $n^4 \equiv -1 \pmod{p}$, then the order can't be 1 or 2, because if the order is 1, then $n^4 \equiv 1$, but then $n^4 \equiv -1 \pmod{p}$ would imply $1 \equiv -1 \pmod{p}$, which would mean $p=2$, but $p$ is prime and $p=2$ would make $n^4 +1 \equiv 0 +1 \equiv 1 \pmod{2}$, which is not divisible by $p^2$. So $p=2$ is not possible. Similarly, if order is 2, then $n^2 \equiv 1 \pmod{p}$, which would imply $n^4 \equiv 1 \pmod{p}$, contradicting $n^4 \equiv -1 \pmod{p}$. So order can't be 1 or 2. If order is 4, then $n^4 \equiv 1 \pmod{p}$, which again would imply $n^4 \equiv 1 \pmod{p}$, so same problem. Therefore, order must be 8. So primes $p$ for which 8 divides the order of $n$ modulo $p$. So $p$ must be such that 8 divides the multiplicative order of $n$ modulo $p$. Wait, maybe not. Wait, if $n$ has order 8 modulo $p$, then the multiplicative group modulo $p$ is cyclic of order $p-1$, so 8 must divide $p-1$. So primes $p$ where 8 divides $p-1$. So $p \equiv 1 \pmod{8}$. But maybe not all primes $p \equiv 1 \pmod{8}$ have this property. Wait, but the problem says "there exists a positive integer $n$ such that $p^2$ divides $n^4 +1$". So perhaps the minimal prime $p$ is 17, since 17 is the smallest prime $\equiv 1 \pmod{8}$. Wait, but 17 is a prime, so maybe 17 is the minimal $p$. Wait, but let me check. Wait, let's see. Let's check for small primes. Let's start with $p=2$. For $p=2$, $n^4 +1 \pmod{2}$. If $n$ is even, then $n^4$ is $0 \pmod{2}$, so $n^4 +1 \equiv 1 \pmod{2}$, which is not divisible by 4. If $p=2$, then $n^4 +1 \equiv 0 \pmod{2}$ would require $n$ even, so $n=2k$. Then $(2k)^4 +1=16k^4 +1$, which is odd, so it's $1 \pmod{16}$. So 16 divides $n^4 +1$? Let's check $n=1$: $1^4 +1=2$, which is not divisible by 16. $n=2$: $16+1=17$, not divisible by 16. $n=3$: $81 +1=82$, not divisible by 16. $n=4$: $256 +1=257$, which is $1 \pmod{16}$. So $p=2$ is not a prime.  Wait, maybe $p=2$ is not the case. 
}
\\
\defaultlightblue{
Wait, but the problem says "positive integer n" so n=1,2,3,..., so $n=1$: $1^4 +1=2$, not divisible by $p=2$ squared=4. $n=2$: 17, not divisible by 4. $n=3$: 82, not divisible by 4. $n=4$: 257, which is $1 \pmod{4}$. So $p=2$ is not a prime. So maybe the first prime is $p=5$. Let's check for $p=5$. If $p=5$, then $n^4 +1 \equiv 0 \pmod{25}$. So $n^4 \equiv -1 \pmod{5}$. Let's check $n=1$: $1^4=1 \equiv 1 \pmod{5}$. $n=2$: $16 \equiv 1 \pmod{5}$. $n=3$: $81 \equiv 1 \pmod{5}$. $n=4$: $256 \equiv 1 \pmod{5}$. So none of them are $\equiv -1 \pmod{5}$. So $p=5$ is not a prime. Wait, maybe $p=13$. Let's check for $p=13$. Let's see if there exists $n$ such that $n^4 \equiv -1 \pmod{13}$. Let's check n=1:1, n=2:$16 \equiv 3$, n=3:$81 \equiv 3 \pmod{13}$, n=4:$256 \equiv -1 \pmod{13}$. Wait, 256 divided by 13 is 19*13=247, remainder 256-247=9. Wait, 256-13*19=256-247=9. So $256 \equiv 9 \pmod{13}$. Wait, 256 divided by 13 is 19*13=247, so 256-247=9. So $256 \equiv 9 \pmod{13}$. So n=4: $4^4=256 \equiv 9 \pmod{13}$. So $9 \equiv -1 \pmod{13}$? $9 \equiv -1 \pmod{13}$? $9 \pmod{13}$ is 9, so no. Wait, 9 is not $\equiv -1 \pmod{13}$. So n=4: $256 \equiv 9 \pmod{13}$. So n=4: $256 \equiv 9 \pmod{13}$, which is 9. So $n^4 \equiv 9 \pmod{13}$, which is not -1. So $p=13$ is not a prime. Wait, maybe $p=17$. Let's check for $p=17$. Let's see if there exists $n$ such that $n^4 \equiv -1 \pmod{17}$. 
}

\textbf{[The logic goes wrong here: Calculation error. $256 \equiv 1 \pmod{17}$, not $-1$. The model then contradicts itself and gets stuck instead of testing other values like $n=8$.]}

\defaultlightblue{
Let's check n=1:1, n=2:16, n=3:$81 \equiv 13 \pmod{17}$, n=4:$256 \equiv -1 \pmod{17}$. Wait, 256 divided by 17 is 15*17=255, so 256-255=1, so $256 \equiv 1 \pmod{17}$? Wait, 256-15*17=255 is 15*17=255, so 256-255=1. So $256 \equiv 1 \pmod{17}$. 

So n=4: $4^4=256 \equiv 1 \pmod{17}$. So $1 \equiv -1 \pmod{17}$? No, $1 \equiv 1 \pmod{17}$. So n=4: $256 \equiv 1 \pmod{17}$. So n=4: $256 \equiv 1 \pmod{17}$. So n=4: $256 \equiv 1 \pmod{17}$. 

... 
}
\\
\defaultlightblue{
So n=4: $256 \equiv 1 \pmod{17}$. So n=4: 256.

The least prime number $p$ for which there exists a positive integer $n$ such that $p^2$ divides $n^4 + 1$ is $\boxed{17}$. The least positive integer $m$ such that $m^4 + 1$ is divisible by $17^2$ is $\boxed{17}$.
}
\\
\defaultblue{

Okay, so the problem is asking for the least prime number $p$ such that there exists a positive integer $n$ where $p^2$ divides $n^4 + 1$. Then, once we find that prime $p$, we need to find the smallest positive integer $m$ such that $m^4 + 1$ is divisible by $p^2$.

First, let's recall that if $p$ divides $n^4 + 1$, then $n^4 \equiv -1 \pmod{p}$. Squaring both sides gives $n^8 \equiv 1 \pmod{p}$. Therefore, the order of $n$ modulo $p$ must divide 8. So, the possible orders are divisors of 8: 1, 2, 4, or 8. For $n^4 \equiv -1 \pmod{p}$, the order must be exactly 8. Therefore, 8 must divide the multiplicative order of $n$ modulo $p$. This implies that the multiplicative group modulo $p$ has an element of order 8, so 8 divides $p - 1$. Therefore, $p \equiv 1 \pmod{8}$.

The smallest prime $p$ congruent to 1 modulo 8 is 17. Let's check if there exists an $n$ such that $17^2$ divides $n^4 + 1$.

We need to find the smallest $m$ such that $17^2$ divides $m^4 + 1$. Let's check small values of $m$:

For $m = 1$: $1^4 + 1 = 2$, not divisible by 17.

For $m = 2$: $16 + 1 = 17$, not divisible by 172.

For $m = 3$: $81 + 1 = 82$, not divisible by 172.

For $m = 4$: $256 + 1 = 257$, which is $1 \pmod{17}$. Not divisible by 172.

For $m = 5$: $625 + 1 = 626$, which is $14 \pmod{17}$. Not divisible by 172.

For $m = 6$: $1296 + 1 = 1297$, which is $1 \pmod{17}$. Not divisible by 172.

For $m = 7$: $2401 + 1 = 2402$, which is $5 \pmod{17}$. Not divisible by 172.

For $m = 8$: $4096 + 1 = 4097$, which is $0 \pmod{17}$. Wait, $17 \times 241 = 4097$. 4097 divided by 17 is 241. So $4097$ is $0 \pmod{17}$. Therefore, 17 divides $4096 + 1$, but we need $17^2$. So 17 divides 4097, but $17^2 = 289$. Let's check 4097 divided by 289: $289 \times 14 = 4046$, so $4097 - 4046 = 51$. $51$ is $3 \times 17$, so $4097 = 17 \times 241$, which is $17 \times 17 \times 14$. Wait, $17 \times 241$ is 4097. So 17 divides 4097, but $17^2$ does not. Therefore, $m=8$: 4097 is $17 \times 241$, which is $17 \times 17 \times 14$, so 17 divides it once, but not squared. Therefore, $m=8$ is not sufficient.
}
\\
\defaultblue{

Wait, but 17 divides 4097, which is $8^4 +1$. So $8^4 +1 = 4097$, which is $17 \times 241$. So 17 divides 4097, but $17^2$ does not. Therefore, $m=8$ is not the answer. Wait, but the problem says "divisible by $p^2$", so we need $m^4 +1$ to be divisible by $17^2$. So $m=8$: 4097 is $17 \times 241$, which is $17 \times 17 \times 14$, so 17 divides it once. Therefore, $m=8$ is not sufficient. Therefore, $m=8$ is not the answer.

Wait, but maybe I need to check higher $m$. Let's check $m=17$: $17^4 +1$. Let's compute $17^4$:

$17^2 = 289$, so $17^4 = 289^2 = (289)^2$. $289^2 = 83521$. $83521 \pmod{17}$: $17 \times 4912 = 83504$, so $83521 - 83504 = 17$. Therefore, 17 divides $83521 - 17 = 83504$. Therefore, 17 divides $17^4 -17$, so 17 divides $17^4 -17$. Therefore, 17 divides $17^4 +1$? Wait, $17^4 +1$. Let's compute $17^4$:

$17^2 = 289$, so $17^4 = 289^2 = 83521$. $83521 +1 = 83522$. 83522 divided by 17: $17 \times 4912 = 83504$. $83522 - 83504 = 18$. So $17 \times 4912 = 83504$, so $83522 - 83504 = 18$. Therefore, 17 divides $83521 - 17 = 83504$. Therefore, 17 divides $83521 -17$, so 17 divides $83521 -17$. Therefore, 17 divides $83521 -17$. Therefore, 17 divides $83521 -17$. Therefore, 17 divides $83521 -17$. Therefore, 17 divides $83521 -17$. Therefore, 17 divides $83521 -17$. Therefore, 17 divides $83521 -17$. Therefore, 17 divides $83521 -17$. Therefore, 17 divides $83521 -17$. Therefore, 17 divides $83521 -17$. 

... 

Therefore, 17 divides $83521 -17$. Therefore, 17 divides $83521 -17$. Therefore, 17 divides $83521 -17$. Therefore, 17 divides $83521 -17$. Therefore, 17 divides $83521 -17$. Therefore, 17 divides $83521 -17$. Therefore, 17 divides $83521 -1$
}
\\
\bottomrule
\label{tab:s1.1_aime24}
\end{longtable}


\end{document}